\documentclass[journal,10pt]{IEEEtran}
\usepackage{graphicx}
\usepackage{amsmath}
\usepackage{caption}
\usepackage{subcaption}
\usepackage{amsfonts}
\usepackage{amssymb}
\usepackage{algorithm,algorithmic}
\usepackage{cite}
\usepackage{array}
\usepackage{soul,color}

\begin{document}

\title{Task-Oriented Prediction and Communication Co-Design for Haptic Communications}

\author{Burak~Kizilkaya,~\IEEEmembership{Graduate Student Member,~IEEE,}
        Changyang~She,~\IEEEmembership{Member,~IEEE,}\\
        Guodong~Zhao,~\IEEEmembership{Senior Member,~IEEE,}
        and~Muhammad~Ali~Imran,~\IEEEmembership{Senior Member,~IEEE}% <-this % stops a space
\thanks{Copyright (c) 2015 IEEE. Personal use of this material is permitted. However, permission to use this material for any other purposes must be obtained from the IEEE by sending a request to pubs-permissions@ieee.org.}% <-this % stops a space
\thanks{B. Kizilkaya, G. Zhao and M. A. Imran are with School of Engineering, University of Glasgow, Glasgow, G12 8QQ, U.K. (e-mail: b.kizilkaya.1@research.gla.ac.uk; guodong.zhao@glasgow.ac.uk; muhammad.imran@glasgow.ac.uk).}% <-this % stops a space
\thanks{C. She is with School of Electrical and Information Engineering, The University of Sydney, Sydney, NSW 2006, Australia. (e-mail: shechangyang@gmail.com).}% <-this % stops a space
}

\maketitle

\begin{abstract}
    
     Prediction has recently been considered as a promising approach to meet low-latency and high-reliability requirements in long-distance haptic communications. However,  most of the existing methods did not take features of tasks and the relationship between prediction and communication into account. In this paper, we propose a task-oriented prediction and communication co-design framework, where the reliability of the system depends on prediction errors and packet losses in communications. The goal is to minimize the required radio resources subject to the low-latency and high-reliability requirements of various tasks. Specifically, we consider the \emph{just noticeable difference} (JND) as a performance metric for the haptic communication system. We collect experiment data from a real-world teleoperation testbed and use \emph{time-series generative adversarial networks} (TimeGAN) to generate a large amount of synthetic data. This allows us to obtain the relationship between the JND threshold, prediction horizon, and the overall reliability including communication reliability and prediction reliability. We take 5G New Radio as an example to demonstrate the proposed framework and optimize bandwidth allocation and data rates of devices. Our numerical and experimental results show that the proposed framework can reduce wireless resource consumption up to $77.80\%$ compared with a task-agnostic benchmark.

\end{abstract}

\begin{IEEEkeywords}
   Resource management, communication system reliability, prediction methods
\end{IEEEkeywords} 

\section{Introduction}\label{introduction}
Haptic communications will lie the foundation for emerging teleoperation applications such as remote surgery and diagnosis in healthcare \cite{latif20175g}, remote laboratory and training in education \cite{kizilkaya20215g}, remote driving in transportation \cite{zhang2020toward}, and advanced manufacturing in Industry 4.0 \cite{gupta2019tactile}. A typical teleoperation system with haptic communications consists of three main domains \cite{aijaz2018toward,antonakoglou2018toward,simsek20165g}, namely master domain, slave domain and communication domain. At master domain, a master device equipped with a haptic interface transmits control commands over a communication system, where the control commands are usually a series of positions, orientations, velocities, or poses. Then, a slave device, in slave domain, executes the received control commands to complete a task and provides haptic feedback over a communication system. To provide high quality user experience, haptic communications need to meet stringent requirements on latency and reliability \cite{antonakoglou2018toward}. These requirements are well-aligned with \emph{ultra-reliable low-latency communications} (URLLC) in the fifth generation (5G) cellular networks \cite{3GPP2017NR}. Specifically, the \emph{end-to-end} (E2E) delay should be around $1$~ms and reliability should be higher than $99.999$\%. However, it is very challenging to meet the latency and reliability requirements, especially when the communication distance between the master and slave devices is longer than $300$~km \cite{simsek20165g}.

In the existing literature \cite{tong2018minimizing, hosseini2016predictive, boabang2020framework, girgis2021predictive, hou2019prediction, wang2021prediction, zhang2021minimizing}, prediction has been considered as a promising approach to meet the stringent \emph{quality of service} (QoS) requirements, such as latency and reliability. In a packetized predictive control system, predicted packets are sent to the receiver in advance and are used in case of packet loss \cite{tong2018minimizing}. The authors of this work jointly optimized prediction and communication systems to minimize wireless resource consumption. In \cite{hosseini2016predictive}, a two-stage prediction framework was proposed for haptic feedback prediction in remote driving to assist the driver in poor network conditions, where prediction errors were not considered \cite{tong2018minimizing,hosseini2016predictive}. In \cite{boabang2020framework}, a haptic feedback prediction framework was proposed in remote surgery use case. Gaussian mixture regression model was used to predict haptic feedback to surgeon in a needle insertion process, where the prediction accuracy was measured by \emph{Root Mean Square Error} (RMSE) between predicted force and ground truth. In \cite{girgis2021predictive}, authors proposed a predictive actuation framework by predicting missing state at receiver using previously received state. A model-based prediction algorithm was used to minimize average \emph{age of information} (AoI) and transmit power. However, in both \cite{boabang2020framework} and \cite{girgis2021predictive}, human perception capabilities were not taken into account.  

Recently, user experience was considered in haptic communications, and was characterized by \emph{just noticeable difference} (JND) \cite{feyzabadi2013human}. Here,  JND is defined as the minimum difference between  the  master  and  the  slave that can be perceived by human users. Since human can only notice the difference when it is larger than a certain JND threshold, it is reasonable to design the predictor by considering the JND violation probability, which is defined as the probability that the difference between the predicted value and the ground truth is larger than a threshold. In \cite{hou2019prediction}, the authors used the JND violation probability as the performance metric to measure the reliability of the predictor and proposed a prediction and communication co-design framework to reduce user experienced delay, where a model-based prediction algorithm was deployed at the transmitter. It was demonstrated that prediction and communication co-design improves the tradeoff between user experienced delay and reliability compared to the communication system with no prediction. In \cite{wang2021prediction}, JND violation probability was used as the overall performance metric to illustrate the benefits of prediction on AoI in \emph{Intelligent Transport Systems} (ITS).  Similarly in \cite{zhang2021minimizing}, the tradeoff between prediction length and AoI was analyzed by using JND violation probability in prediction design. The results showed that the prediction algorithm used in the system can help to improve the AoI performance.

In practice, the JND threshold is a task dependent parameter with large variation. For example, the JND threshold value would be different when human operator (master) controls a robotic arm (slave) for different tasks. The large JND threshold is expected when human operator has the full arm movement. In contrast, the small JND threshold is required in fine control. However, current design methods cannot capture such JND threshold dynamics since the fundamental relationship between JND threshold and QoS is not clear yet. As a result,  current design methods suffer from over-provisioning of wireless resource that impedes their implementation in real-world systems. Recently proposed task-oriented (or, goal-oriented/semantic) communications \cite{strinati20216g, uysal2021semantic, mostaani2021task} advocates similar approach showing that overall system performance not only depends on bit-level performance but also whether the intended task is accomplished or not, given the application scenario and communication resources available. Such a perspective envisions a paradigm shift in communication system design from bit-level to task-level. However, this is not an easy task since task-oriented communications is the combination of different principles from control theory, information theory and computer science. Effectiveness and efficiency can be enhanced by jointly considering different task objectives and task dependent parameters. In this study, we consider JND as a task dependent parameter and jointly design prediction and communication systems.

In addition, it is very challenging to obtain the relationship between JND threshold and QoS. Unlike communication systems that are built upon fundamental theories, most of the prediction algorithms are developed via data-driven design. This raises a huge challenge when the required error probability is at the level of $10^{-5}$, which requires a large number of real-world data samples to evaluate the probability of the rare event. If we use model-based prediction, it is possible to derive the prediction error probability, but the mismatch between the over simplified theoretical models and practical complex systems will lead to inaccurate results. Therefore, we need to develop innovative methods to overcome the above challenge. 

In this paper, we propose a task-oriented prediction and communication co-design framework in the context of teleoperation system, where the utilization efficiency of the communication system is maximized subject to the requirements of different operating tasks. In particular, in the scenario with limited real-world data samples, we generate synthetic data via time-series generative adversarial networks (TimeGAN). This allows us to obtain the relationship between JND threshold and the prediction error probability that is below $10^{-5}$. We take 5G New Radio as an example to demonstrate the proposed framework, where we compare the performance of the systems with and without task-oriented design. We further compare the performance difference of deploying the predictor at transmitter and receiver sides. Specifically, the main contributions of this paper are listed below:
\begin{itemize}
    \item We propose a task-oriented prediction and communication co-design framework with a predictor at the receiver side. We derived an upper bound of the overall error probability in the framework by taking packet losses and prediction errors into account. From the upper bound, we reveal the tradeoff between the resource utilization efficiency and the overall reliability.
    \item To illustrate how to use this framework in practical system design, we take remote robotic control in 5G New Radio as an example. Then, we formulate an optimization problem to optimize bandwidth allocation and communication data rate subject to constraints on the E2E delay and overall reliability. An optimization algorithm is proposed to find the optimal solution.
    \item We collect the real-world data from a teleoperation prototype. We further use TimeGAN to generate synthetic data for predictor training and testing. With both synthetic data and real-world data, we illustrate the tradeoff between the prediction horizon and prediction error probability, and further evaluate the overall reliability. Our results show that the proposed task-oriented prediction framework can save up to $77.80\%$ bandwidth compared with a benchmark design that is task-agnostic. 
\end{itemize}

\noindent The rest of the paper is organized as follows. In Section \ref{system model}, we develop a general design framework for task-oriented prediction and communication co-design. In Section \ref{tradeoff_analyses}, we illustrate packet losses in communications by taking 5G New Radio as an example. In Section \ref{prediction models}, we introduce prediction algorithms. In Section \ref{resource allocation}, we propose efficient resource allocation with task-oriented prediction. In Section \ref{evaluation}, we present simulation and numerical results. Section \ref{conclusion} concludes this paper. Notations used throughout the paper are listed in Table \ref{tab:notations} for clarification.

\begin{table}
  \caption{Descriptions of Notations}
    \centering
    \begin{tabular}{|p{1cm}|p{6.8cm}|}
    \hline
        \textbf{Notation} & \textbf{Description} \\
        \hline
        $K_m(t)$ & control command send by $m$-th transmitter at time slot $t$  \\
        \hline
        $n$ & number of features in a control command \\
        \hline
        $D^{\rm c}_m$ & E2E communication delay for the $m$-th link \\
        \hline
        $D^{\rm e}_m$ & user experienced delay for the $m$-th link \\
         \hline
        $D^{\rm r}_m$ & core network and backhaul delay for the $m$-th link  \\
         \hline
        $D^{\rm q}_m$ & queuing delay for the $m$-th link \\
        \hline
        $D^{\rm t}_m$ & transmission delay for the $m$-th link  \\
        \hline
        $D^{\rm ch}_m$ & coherence time for the $m$-th link  \\
        \hline
        $D^{\rm th}_m$ & queuing delay threshold for the $m$-th link  \\
        \hline
        $T^\text{p}_m$ & prediction horizon for the $m$-th task  \\
        \hline
        $T_{\rm th}$ & prediction horizon threshold  \\
        \hline
        $D^{\max}_m$ & delay requirement for the $m$-th task  \\
        \hline
        $\epsilon^\text{q}_m$ & queuing delay bound violation probability for the $m$-th task  \\
        \hline
        $\epsilon^\text{d}_m$ & decoding error probability for the $m$-th task  \\
        \hline
        $\epsilon^\text{p}_m$ & prediction error probability for the $m$-th task \\
        \hline
        $\epsilon^{\rm o}_m$ & overall error probability for the $m$-th task  \\
        \hline
        $\epsilon_m^{\max}$ & maximum tolerable error probability for the $m$-th task  \\
        \hline
        $\delta_m$ & JND threshold for the $m$-th task \\ 
        \hline
        $\lambda_m$ & average packet arrival rate for the $m$-th link \\
        \hline
        $E^\text{B}_m$ & effective bandwidth for the $m$-th link \\
        \hline
        $b_m$ & number of bits per packet for the $m$-th link \\
        \hline
        $W_m$ & bandwidth for the $m$-th link \\
        \hline
        $C_m$ & capacity for the $m$-th link \\
        \hline
        $R_m$ & achievable rate for the $m$-th link \\
        \hline
        $V_m$ & channel dispersion for the $m$-th link \\
        \hline
        $\gamma_m$ & SNR for the $m$-th link \\
        \hline
        $\alpha_m$ & large-scale gain for the $m$-th link \\
        \hline
        $g_m$ & small-scale gain for the $m$-th link \\
        \hline
        $P_m$ & maximum transmit power for the $m$-th link \\
        \hline
        $N_0$ & single sided noise spectral density \\
        \hline
        $l_m$ & blocklength for the $m$-th link \\
        \hline
        $A$ & number of critical tasks \\
        \hline 
        $B$ & number of non-critical tasks \\
        \hline
        $Q(.)$ & Q-function \\
        \hline
        $Q^{-1}(.)$ & inverse of the Q-function \\
        \hline
        $\text{W}_{-1}(.)$ & -1 branch of Lambert W-function \\
        \hline
        $f_{\epsilon_m^{\rm p}}(.)$ & function of prediction error probability for the $m$-th task\\
        \hline
        $f_{\epsilon_m^{\rm q}}(.)$ & function of queuing delay bound violation probability for the $m$-th task\\
        \hline
    \end{tabular}
    \label{tab:notations}
\end{table}

\section{A General Design Framework}\label{system model}

\begin{figure*}
    \centering
    \includegraphics[width=0.9\textwidth]{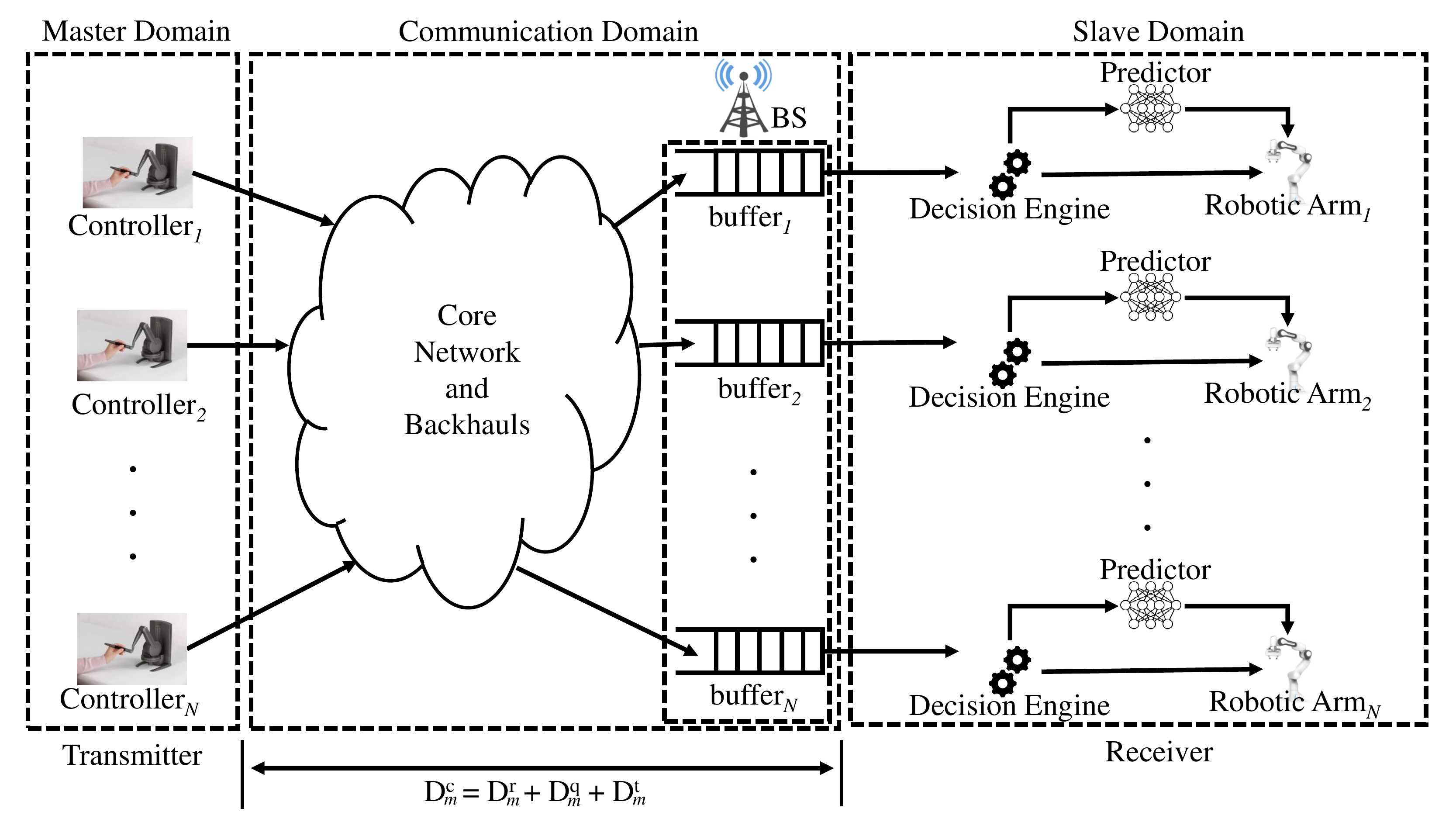}
    \caption{Proposed task-oriented prediction and communication co-design framework system model with considered haptic communication system, where a human can remotely control robots via a communication system. The predictors are deployed at receivers to reduce user experienced delays.}
    \label{fig:system_model}
\end{figure*}

We consider a haptic communication system shown in Fig. \ref{fig:system_model}, where a human can remotely control robots via a communication system. $N$ pairs of transmitters and receivers are considered over a shared wireless channel, where orthogonal subchannels are assigned to different transceiver pairs to avoid interference. The predictors are deployed at receivers to reduce user experienced delays. It is worth noting that the transmitter and receiver in the proposed design framework might be any device depending on the application scenario. As an example of a mission-critical application, a teleoperation scenario in haptic communications is considered in this study. In a tele-driving \cite{kim2022feasibility} scenario, the receiver can be a remote vehicle, while the transmitter is the remote controller. Similarly, in an edge-assisted autonomous driving application \cite{li2021adaptive}, the receiver can be an autonomous vehicle with the control unit on the edge as a transmitter.

\subsection{User Experienced Delay and Delay Requirement}

Time is discretized into slots. At the $t$-th time slot, the $m$-th transmitter sends the control command $K_m(t) = [k^1_m(t), k^2_m(t),..., k^n_m(t)]^T$ to the receiver, where $n$ is the number of features in the command (e.g., joint angles, angular velocity, forces, and torque). Then, $K_m(t)$ arrives at the $m$-th receiver at $(t+D^{\rm c}_m)$-th time slot, where $D^{\rm c}_m$ is E2E communication delay. If the communication system has no prediction capability, the time delay experienced by the user, denoted as $D^{\rm e}_m$, is the same as the communication delay, i.e., $D^{\rm e}_m = D^{\rm c}_m$\footnote{There are different delay components such as computing delay and control delay. However, we focus on communication system design in this work and ignore other delay components.}. Here, the  communication delay $D^{\rm c}_m$ consists of the delay in core network and backhauls $D^{\rm r}_m$, the queuing delay in the buffer of the base station (BS) $D^{\rm q}_m$, and the transmission delay in the radio access network $D^{\rm t}_m$. Then, we have $D^{\rm c}_m = D^{\rm r}_m + D^{\rm q}_m+D^{\rm t}_m$.

In this work, we consider a block fading channel. Given the 3GPP specifications\cite{3GPP2017NR}, $D^{\rm r}_m$ and $D^{\rm t}_m$ are assumed to be bounded by a constant, but $D^{\rm q}_m$ is a random variable. As a result, the communication delay $D^{\rm c}_m$ is a random variable that could be longer than the maximum tolerable delay bound $D^{\max}_m$. To reduce the user experienced delay, $D^{\rm e}_m$, which is defined as the time difference between the communication delay, $D^{\rm c}_m$, and the prediction horizon, $T^{\rm p}_m$, i.e., $D^{\rm e}_m = D^{\rm c}_m - T^{\rm p}_m$, a predictor is equipped at each receiver to predict the delayed or lost trajectories. Given the definition of the user experienced delay, predicting lost or delayed packets reduces user experienced delay by the amount of the prediction horizon, $T^{\rm p}_m$.  Here, lost trajectories are the consecutive packets that cannot be decoded by the receiver due to deep fading in a block fading channel. We denote the maximum prediction horizon $T_{\rm th}$ which is the upper bound of the prediction horizon $T^{\rm p}_m$. This means that the predictor cannot predict the trajectory beyond the maximum prediction horizon $T_{\rm th}$ because the temporal correlation of data becomes very weak. 

Since the user experience delay is determined by the relationship among the communication delay $D_m^c$ and the prediction horizon $T^{\rm p}_m$, we need to consider the following three cases:
\begin{itemize}
    \item Case 1: $D^{\rm c}_m \in (0,D^{\max}_m]$
    
    The receiver only needs the predicted trajectory when some packets are lost. The prediction horizon depends on the number of consecutive packet losses. We consider a block fading channel. The channel gain remains constant within duration $D^{\rm ch}_m$, and varies independently from one duration to another. $D^{\rm ch}_m$ is referred to as the channel coherence time. When the wireless channel is in deep fading, most of the packets cannot be decoded by the receiver. Given this fact, it is reasonable to assume that $D^{\rm ch}_m$ is upper bound of the time horizon with consecutive packet losses.\footnote{The probability that the channel stays in deep fading in multiple consecutive blocks is extremely small, i.e., much smaller than $10^{-5}$, and hence is not considered.} With the block fading channel, the prediction horizon is bounded by $T^{\rm p}_m \leq D^{\rm ch}_m$. In this case, the user experienced delay is equal to the communication delay, i.e., 
    \begin{equation}\nonumber
      D^\text{e}_m = D^\text{c}_m \leq D^{\max}_m.
    \end{equation}

    \item Case 2: $D^{\rm c}_m \in (D^{\max}_m, T_{\rm th}+D^{\max}_m]$
    
    To satisfy the delay requirement, the prediction horizon is $T^\text{p}_m = D^\text{c}_m - D^{\max}_m$.  With the help of the predictor, the user experienced delay becomes 
    \begin{equation}\nonumber
       D^{\rm e}_m = D^{\rm c}_m - T^{\rm p}_m = D^{\max}_m.
    \end{equation}

    \item Case 3: $D^{\rm c}_m \in (T_{\rm th}+D^{\max}_m,\infty)$
    
    In this case, it is not possible to meet the delay requirement even with the maximum prediction horizon. Then, the user experienced delay becomes
    \begin{equation}\nonumber
        D^{\rm e}_m = D^{\rm c}_m - T^{\rm p}_m > D^{\max}_m.
    \end{equation}
    
\end{itemize}

For haptic communications, the delay experienced by a user should not exceed a delay bound. The maximum tolerable delay bound of the $m$-th user is denoted by $D^{\max}_m$. Then, the user experienced delay, $D^{\rm e}_m$, should satisfy the following requirement
\begin{equation}
    D^{\rm e}_m \leq D^{\max}_m.
    \label{delay_req}
\end{equation}
It is worth noting that the above constraint cannot be satisfied with probability one since the queuing delay is stochastic in most of communication systems. In the next subsection, we will analyze the delay bound violation probability.

\subsection{Reliability Components and Reliability Requirement}
Let's denote the overall error probability and the maximum tolerable error probability of the $m$-th task by $\epsilon^{\rm o}_m$ and $\epsilon_m^{\max}$, respectively. The overall error probability consists of decoding error probability, $\epsilon^\text{d}_m$, queuing delay bound violation probability, $\epsilon^\text{q}_m$, and prediction error probability, $\epsilon^\text{p}_m$.  

The decoding error probability of the $m$-th user depends on the channel fading, the resource allocation policy, and the modulation and coding scheme in wireless communications. We need to optimize the communication system to obtain a satisfactory decoding error probability. In the next section, we will provide the expression of $\epsilon^\text{d}_m$ in a specific communication system.

The prediction error probability is defined as the probability that the prediction error is larger than the required JND threshold of a task. For the $m$-th user, the JND threshold is denoted by $\delta_m$. For a given prediction algorithm, the relationship between prediction error probability, $\epsilon^\text{p}_m$, the prediction horizon, $T^\text{p}_m$, and the JND threshold is characterized by the following function,
\begin{align}
   \label{eq:perror}
   f_{\epsilon^\text{p}_m}(T^\text{p}_m,\delta_m) = & \Pr\{|{\hat{K}_m(t+T^\text{p}_m)} - K_m(t+T^\text{p}_m)|  > \delta_m\}, \\ \nonumber & t=1,2,3,... 
\end{align}

where $\hat{K}_m(t+T^\text{p}_m)$ is the predicted trajectory for the $(t+T^\text{p}_m)$-th slot and ${K_m(t+T^\text{p}_m)}$ is the actual trajectory in this slot. As shown in \cite{hou2019prediction}, $f_{\epsilon^\text{p}_m}(T^\text{p}_m,\delta_m)$ increases with $T^\text{p}_m$ and decreases with $\delta_m$.

Similarly, we denote the relationship between the queuing delay bound and the delay bound violation probability by a function $f_{\epsilon^\text{q}_m}(\kappa)$. It is the probability that the queuing delay, $D^\text{q}_m$, is greater than a queuing delay bound, $\kappa$, i.e.,
\begin{equation}
   f_{\epsilon^\text{q}_m}(\kappa) = \Pr\{D^\text{q}_m > \kappa\}. 
\end{equation} 
Since the delay bound violation probability decreases with the required delay bound, $f_{\epsilon^\text{q}_m}(\kappa)$ is a monotonic decreasing function. To meet the maximum delay bound, a threshold of queuing delay is given by $D^{\rm th}_m = D^{\max}_m - D^{\rm t}_m - D^{\rm r}_m$. 

In the sequel, we analyze the overall error probability in the three cases discussed in the previous subsection.

\begin{itemize}
    \item Case 1: $D^{\rm c}_m \in (0,D^{\max}_m]$
    
    In this case, we have $D^\text{q}_m \leq D^{\rm th}_m$. From the definition of $f_{\epsilon_m^{\rm q}}(\kappa)$, the probability that the queuing delay does not exceed $D_m^{\rm th}$ can be expressed by substituting $D_m^{\rm th}$ into $\kappa$, so that $\kappa = D_m^{\rm th}$ and the probability that the queuing delay does not exceed $D_m^{\rm th}$ becomes
    \begin{align}
        \Pr\{D_m^\text{q}\leq D_m^{\rm th}\} = 1-f_{\epsilon_m^{\rm q}}(D_m^{\rm th}).
    \end{align}

    Since the maximum prediction horizon does not exceed $D^{\rm ch}_m$, the prediction error probability is bounded by $f_{\epsilon_m^{\rm p}}(D_m^{\rm ch},\delta_m)$. Given the decoding error probability, $\epsilon_m^{\rm d}$, the error probability in case 1 can be expressed as $\epsilon^\text{c,1}_m = f_{\epsilon^\text{p}_m}(D_m^{\rm ch},\delta_m) \epsilon^\text{d}_m$.

    \item Case 2: $D^{\rm c}_m \in (D^{\max}_m, T_{\rm th}+D^{\max}_m]$
    
    In this case, $D^{\rm q}_m \in (D_m^{\rm th},D_m^{\rm th}+T_{\rm th}]$, which happens with a probability of
    \begin{align}
        &\Pr\{D^{\rm q}_m \in (D_m^{\rm th},D_m^{\rm th}+T_{\rm th}]\} =\nonumber\\ &f_{\epsilon_m^{\rm q}}(D_m^{\rm th}) - f_{\epsilon_m^{\rm q}}(D_m^{\rm th}+T_{\rm th}).
    \end{align}
    The prediction horizon is bounded by $T_{\rm th}$, and thus the error probability is bounded by $\epsilon_m^{\rm c,2} \leq f_{\epsilon_m^{\rm p}}(T_{\rm th}, \delta_m)$.
    
    \item Case 3: $D^{\rm c}_m \in (T_{\rm th}+D^{\max}_m,\infty)$
    
    In this case, $D^\text{q}_m \in (T_{\rm th}+D_m^{\rm th},\infty)$. Case 3 happens with a probability of 
    \begin{align}
        \Pr\{D^{\rm q}_m \in (T_{\rm th}+D_m^{\rm th},\infty)\} = f_{\epsilon_m^{\rm q}}(D_m^{\rm th}+T_{\rm th}).
    \end{align}
    Since the delay requirement is not satisfied, all the packets are lost. The error probability in this case is $\epsilon_m^{\rm c,3} = 1$.
\end{itemize}
The overall error probability is a combination of error probabilities in the above three cases. It can be expressed as follows,
\begin{align}
       \epsilon^{\rm o}_m  = &\epsilon^{\text{c},1}_m \Pr\{D_m^\text{q}\leq D_m^{\rm th}\} + \epsilon^{\text{c},2}_m \Pr\{D^{\rm q}_m \in (D_m^{\rm th},D_m^{\rm th} + T_{\rm th}]\} \nonumber\\
       &+ \epsilon_m^{\rm c,3} \Pr\{D^{\rm q}_m \in (T_{\rm th}+D_m^{\rm th},\infty)\}\nonumber\\
        \leq & f_{\epsilon^\text{p}_m}(D_m^{\rm ch},\delta_m) \epsilon^\text{d}_m (1 - f_{\epsilon^\text{q}_m}(D_m^{\rm th})) \nonumber\\  
        &+ f_{\epsilon^\text{p}_m}(T_{\rm th},\delta_m) (f_{\epsilon_m^{\rm q}}(D_m^{\rm th}) - f_{\epsilon_m^{\rm q}}(D_m^{\rm th}+T_{\rm th})) \nonumber\\
        &+ f_{\epsilon^\text{q}_m}( D_m^{\rm th}+T_{\rm th}).
    \label{overall_error}
\end{align}
    
The reliability requirement  $\epsilon^{\rm o}_m \leq \epsilon_m^{\max}$ can be satisfied if the upper bound in \eqref{overall_error} meets the following constraint,
\begin{align}
  &f_{\epsilon^\text{p}_m}(D_m^{\rm ch},\delta_m) \epsilon^\text{d}_m (1 - f_{\epsilon^\text{q}_m}(D_m^{\rm th}))  + f_{\epsilon^\text{p}_m}(T_{\rm th},\delta_m) (f_{\epsilon_m^{\rm q}}(D_m^{\rm th}) \nonumber\\
  &- f_{\epsilon_m^{\rm q}}(D_m^{\rm th}+T_{\rm th})) + f_{\epsilon^\text{q}_m}( D_m^{\rm th}+T_{\rm th})  \leq \epsilon_m^{\max}.
   \label{reliability_req}
\end{align}

We denote the general utilization efficiency of a communication system by $U(\bf{{x}})$, where ${\bf{x}} = [x_1,...,x_N]$ is the optimization variables of the $N$ tasks. A general task-oriented prediction and communication co-design framework can be formulated as follows,
\begin{align}
    \max_{\bf{x}}
    & \; {U(\bf{x})} \label{generalObjective}\\
    \text{s.t.} & \;
    \eqref{delay_req}\; \text{and}\; \eqref{reliability_req}.\nonumber
\end{align}

With this framework, we can jointly optimize prediction and communication systems to achieve better resource utilization efficiency.

\section{5G New Radio: An Example of Communication System}\label{tradeoff_analyses}
To illustrate how to obtain the upper bound in (\ref{overall_error}), we consider 5G New Radio as an example in the rest part of this work and derived the decoding error probability, $\epsilon^\text{d}_m$, and the queuing delay bound violation probability, $f_{\epsilon^\text{q}_m}(.)$, in this section.

\subsection{Decoding Error Probability}
To achieve low transmission delay, the blocklength of channel codes is short. In the finite blocklength regime, the maximal achievable rate can be accurately approximated as \cite{yang2014quasi}
\begin{equation}
    R_m  \approx C_m - \sqrt{\frac{V_m}{l_m}} Q^{-1}(\epsilon^\text{d}_m)\; \text{(bits/s/Hz),}
\end{equation}

\noindent where $C_m = \log(1+\gamma_m)$ is the Shannon capacity, $\gamma_m = \frac{\alpha_m g_m P_m}{N_0W_m}$ is the received signal to noise ratio (SNR) at BS, $\alpha_m$ is the large-scale channel gain, $g_m$ denotes the small-scale channel gain, $P_m$ denotes the transmit power, $N_0$ is the single sided noise spectral density, $V_m = \log(e)^2 \left[1 - \frac{1}{(1+\gamma_m)^2}\right]$ is the channel dispersion, $l_m = D^\text{t}_m W_m $ is the blocklength, $D^\text{t}_m$ is the transmission duration, $W_m$ is the bandwidth, and $Q^{-1}(.)$ is the inverse of the Q-function. Then, decoding error probability can be expressed as
\begin{equation}
    \epsilon^\text{d}_m \approx Q\left( \frac{D^\text{t}_m W_m C_m - b_m + \log( D^\text{t}_m W_m)/2}{\sqrt{D^\text{t}_m W_m V_m}}\right),
\end{equation}

\noindent where $b_m = D^\text{t}_m W_m R_m$ is the number of information bits. If SNR is higher than 5dB, channel dispersion $V_m$ becomes $\log(e)^2$ \cite{schiessl2015delay}. Then, $\epsilon^\text{d}_m$ becomes
\begin{equation}
     \epsilon^\text{d}_m \approx Q\left( \frac{D^\text{t}_m W_m C_m - b_m + \log( D^\text{t}_m W_m)/2}{\log(e) \sqrt{D^\text{t}_m W_m}}\right).
     \label{eps_d_formula}
\end{equation}

From (\ref{eps_d_formula}), we can obtain the following Lemma.

\textit{Lemma 1:} Given $W_m$, $\epsilon^\text{d}_m$ increases with $b_m$.

\textit{Proof:} Given $W_m$, the input of the Q-function in (\ref{eps_d_formula}) decreases with $b_m$. Since Q-function is a decreasing function, $\epsilon^\text{d}_m$ increases with increasing $b_m$.

\subsection{Queuing Delay Violation Probability}
In URLLC, the transmission delay (transmission time interval could be $0.125$~ms in 5G) is much shorter than the channel coherence time. Thus, the service rate of the queuing system is a constant. We use effective bandwidth to characterize the minimum service rate that is required to achieve the delay bound and delay bound violation probability \cite{chang1995effective}. To derive the closed-form expression of queuing delay violation probability, we further assume that the packet arrival processes are Poisson processes. Denote the average packet arrival rate of the $m$-th user by $\lambda_m$ with the unit of (packets/s). As discussed in \cite{she2017cross,hou2019prediction}, the queuing delay violation probability decreases exponentially as the delay bound increases, i.e.,
\begin{align}
    & f_{\epsilon^\text{q}_m}(\kappa) = e^{\kappa \xi(E^\text{B}_m,\lambda_m)},
    \label{eps_q_func}\\
    &\xi(E^\text{B}_m,\lambda_m) = E^\text{B}_m \text{W}_{-1}\left(-\frac{\lambda_m}{E^\text{B}_m}e^{-\frac{\lambda_m}{E^\text{B}_m}}\right) + \lambda_m
    \label{eff_band_func},
\end{align}
where $\text{W}_{-1}(.)$ is the -1 branch of Lambert W-function, which is defined as the inverse function of
$f(x) = xe^x$ and $E^\text{B}_m$ is the effective bandwidth, which is the required service rate. To meet the queuing delay bound $\kappa$ and the delay bound violation probability in \eqref{eps_q_func}, the number of bits transmitted in a transmission time interval should satisfy the following expression
\begin{align}
b_m/D^\text{t}_m = E^\text{B}_m. \label{eq:bitTTI}
\end{align}
Upon substituting $E^\text{B}_m$ into
\eqref{eps_q_func} and \eqref{eff_band_func}, the queuing delay bound violation probability can be expressed as
\begin{equation}
    f_{\epsilon^\text{q}_m}(\kappa) = \text{exp}\Bigg\{\kappa \Bigg[\frac{b_m \text{W}_{-1}\left(-\frac{\lambda_m D^\text{t}_m}{b_m}e^{-\frac{\lambda_m D^\text{t}_m}{b_m}}\right)}{D^\text{t}_m} + \lambda_m \Bigg] \Bigg\}.
    \label{eps_q_formula}
\end{equation}

With the expression given in (\ref{eps_q_formula}), the following property of queuing delay bound violation probability function can be obtained. 

\textit{Lemma 2:} For given $\kappa$, $\lambda_m$, and $D^\text{t}_m$, $f_{\epsilon^\text{q}_m}(\kappa)$ strictly decreases with $b_m$.

\textit{Proof:} To check the monotonicity of $f_{\epsilon^\text{q}_m}(\kappa)$ in terms of $b_m$, we have the partial derivative in (\ref{partial_derivative}).

\begin{figure*}
\begin{equation}
\begin{split}
\frac{\partial f_{\epsilon^\text{q}_m}}{\partial b_m} = & \kappa\,\text{exp}\left\{{\kappa\,\left(\lambda_m +\frac{b_m\,{\mathrm{W}}_{-1}\left(-\frac{{D^\text{t}_m}\,\lambda_m \,{\mathrm{e}}^{-\frac{{D^\text{t}_m}\,\lambda_m }{b_m}}}{b_m}\right)}{{D^\text{t}_m}}\right)}\right\} \times \,\left(\frac{{\mathrm{W}}_{-1}\left(-\frac{{D^\text{t}_m}\,\lambda_m \,{\mathrm{e}}^{-\frac{{D^\text{t}_m}\,\lambda_m }{b_m}}}{b_m}\right)}{{D^\text{t}_m}} \right.\\
& \left. -\frac{b_m^2\,{\mathrm{e}}^{\frac{{D^\text{t}_m}\,\lambda_m }{b_m}}\,\left(\frac{{D^\text{t}_m}\,\lambda_m \,{\mathrm{e}}^{-\frac{{D^\text{t}_m}\,\lambda_m }{b_m}}}{b_m^2}-\frac{{{D^\text{t}_m}}^2\,\lambda_m ^2\,{\mathrm{e}}^{-\frac{{D^\text{t}_m}\,\lambda_m }{b_m}}}{b_m^3}\right)\,{\mathrm{W}}_{-1}\left(-\frac{{D^\text{t}_m}\,\lambda_m \,{\mathrm{e}}^{-\frac{{D^\text{t}_m}\,\lambda_m }{b_m}}}{b_m}\right)}{{{D^\text{t}_m}}^2\,\lambda_m \,\left({\mathrm{W}}_{-1}\left(-\frac{{D^\text{t}_m}\,\lambda_m \,{\mathrm{e}}^{-\frac{{D^\text{t}_m}\,\lambda_m }{b_m}}}{b_m}\right)+1\right)}\right)<0. 
\end{split}\label{partial_derivative}
\end{equation} 
\hrulefill
\end{figure*}
Since $W_{-1}(.)$ is always negative, and exponential function is always positive, the partial derivative in (\ref{partial_derivative}) is negative. Therefore, $f_{\epsilon^\text{q}_m}(\kappa)$ decreases with $b_m$ when $\kappa$, $\lambda_m$, and $D^\text{t}_m$ are given.

%%%%%%%%%%%%%%%%%%%%%%%%%%%

\section{TimeGAN Assisted Prediction: An Example of Prediction Algorithm}\label{prediction models}
To characterize the tradeoff between the prediction error probability, the prediction horizon, and the JND threshold, we require a large number of trajectories for evaluating the prediction error probability below $10^{-5}$. \footnote{The reliability requirement is defined by the 5G standard in \cite{3GPP2017NR}.} However, collecting enough trajectories in a real-world robotic platform may take years. To address this issue, we generate synthetic data using TimeGAN. In this section, we provide details of real-world data collection, synthetic data generation, training and testing of different prediction algorithms.

\subsection{Real-world Data Set Collection}\label{subsec:realdata}
\begin{figure}
    \centering
    \includegraphics[width = 0.49\textwidth]{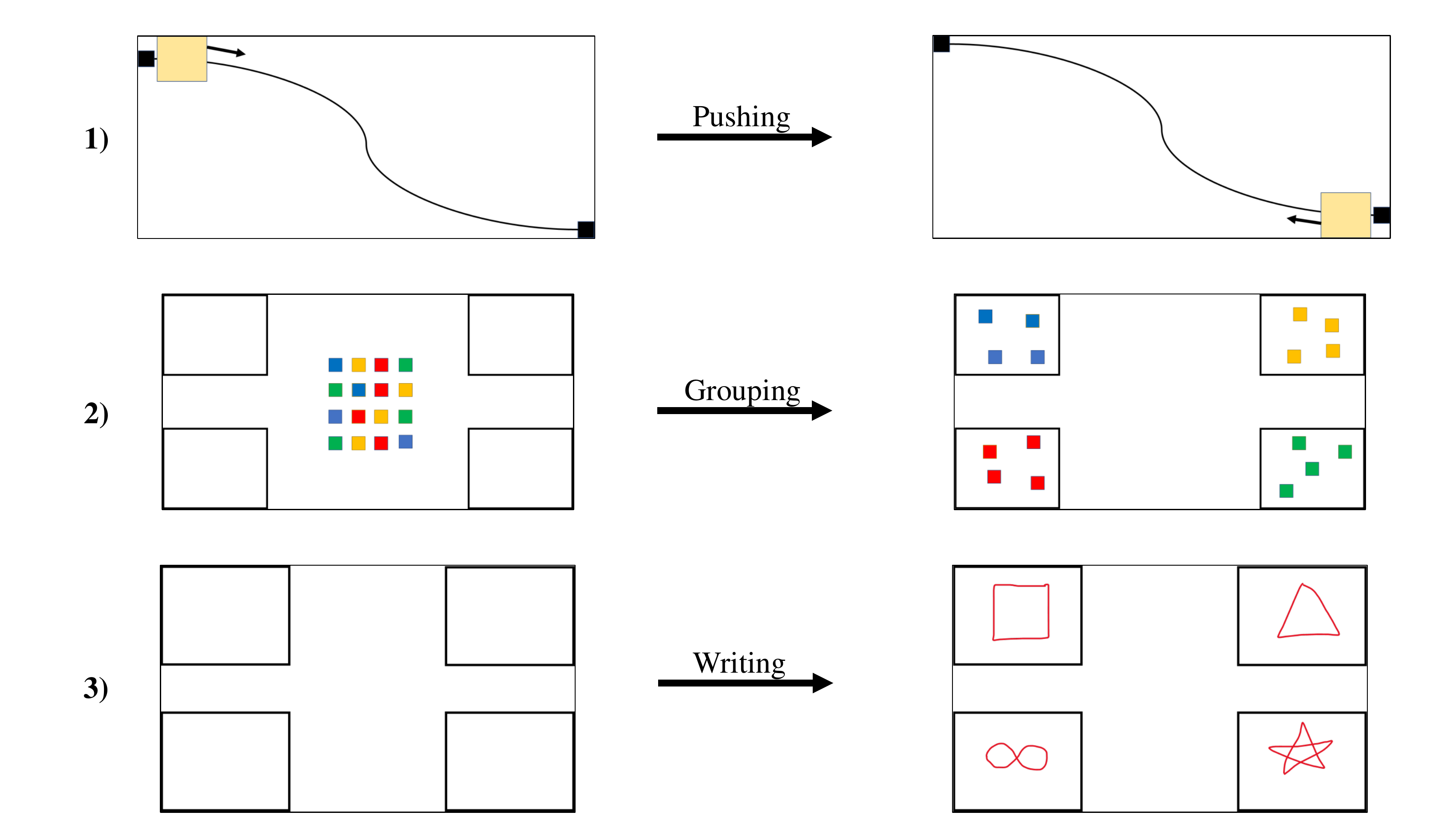}
    \caption{Tasks for data collection.}
    \label{fig:tasks}
\end{figure}

The real-world trajectory data samples\footnote{The dataset is available at: http://dx.doi.org/10.5525/gla.researchdata.1392} are collected from our teleoperation testbed\footnote{The demonstration video: https://youtu.be/c3onK5Vh6QE}. The robotic arm is controlled by a human user to finish three types of tasks as shown in Fig. \ref{fig:tasks}.
\begin{enumerate}
    \item \textit{Pushing a box:} Push a small box from the starting point to the end point along a given routine.
    
    \item \textit{Grouping items with different colors:} Move items with the same color to the same area.
    
    \item \textit{Writing symbols:} Write symbols by controlling the robotic arm.
\end{enumerate}

Each trajectory is a time-series of observations given by $\textbf{k}_{t:t^{'}} = \{\textbf{k}_{t}, \textbf{k}_{t+1},...,\textbf{k}_{t^{'}}\}$ where each observation, $\textbf{k}_t = [q_t,\dot{q_t}]$, consists of an angular position, $q_t$, and angular velocity, $\dot{q_t}$, in the $t$-th time slot. In experiments, we recorded around $1.7\times 10^7$ observations with timestamps at the frequency of one thousand observations per second.

\subsection{Synthetic Data Set Generation}
To generate synthetic data, we apply TimeGAN \cite{yoon2019time}, which is a framework for time-series data generation. TimeGAN framework consists of four neural networks namely embedding network, recovery network, sequence generator, and sequence discriminator. The parameters of the four neural networks are denoted by $\theta_{\rm em}$, $\theta_{\rm re}$, $\theta_{\rm g}$, and $\theta_{\rm d}$, respectively.

Embedding and recovery networks are mappings from feature space to latent space and vice versa. Let's denote the latent variable and the latent space by $\textbf{v}_{i:i'}$ and $\mathcal{V_T}$, respectively. Then, the embedding network outputs a sequence of latent variables from a given sequence of features, i.e., $\textbf{v}_{i:i'} = \vartheta_{\rm em}(\textbf{k}_{i:i'}|\theta_{\rm em})$. The recovery network reconstructs features from latent variables according to $\textbf{\^{k}}_{i:i'} = \vartheta_{\rm re}(\textbf{v}_{i:i'}|\theta_{\rm re})$.

The generator network is denoted by $\textbf{\^{v}}_{i:i'} = \vartheta_{\rm g}(\textbf{z}_{i:i'}|\theta_{\rm g})$. It takes a sequence of random variables $\textbf{z}_{i:i'}$ from a known distribution as its input and generates synthetic sequences in the latent space, $\textbf{\^{v}}_{i:i'}$. The discriminator is a classification network in the latent space. Given a sequence of latent variables, a synthetic sequence or a real sequence, the output of the discriminator is an indicator $\tilde{y} = \vartheta_{\rm d}(\textbf{{v}}_{i:i'}|\theta_{\rm d})$. If $\tilde{y}=1$ the sequence is classified as a real one. Otherwise, the sequence is classified as a synthetic one.

In TimeGAN, the four components are trained jointly with three loss functions: reconstruction loss, $\mathcal{L_\text{R}}$, unsupervised loss, $\mathcal{L_\text{U}}$, and supervised loss, $\mathcal{L_\text{S}}$.

Reconstruction loss is used to measure the difference between the reconstructed features, $\textbf{\^{k}}_{i:i'}$, and the original features, $\textbf{{k}}_{i:i'}$ i.e.,
\begin{align}
    &\mathcal{L_\text{R}} = \mathbb{E}_{\textbf{{k}}_{i:i'}} \left[  \sum_t ||\textbf{{k}}_{t} - \textbf{\^{k}}_{t}||_2 \right]. \label{reconstruction_loss}
\end{align}
Unsupervised loss comes from the zero-sum game as in conventional GAN \cite{creswell2018generative}. It maximizes the likelihood of correct classifications for the discriminator. Supervised loss, on the other hand, is introduced to check the discrepancy between real and synthetic data distributions in latent space. The definitions of $\mathcal{L_\text{U}}$ and $\mathcal{L_\text{S}}$ can be found in \cite{yoon2019time,yoon2019time-supp}.

The generator and discriminator networks are trained iteratively by solving the following problem,
\begin{IEEEeqnarray}{c'l'l'l}
    \IEEEyesnumber\label{timegan_Problem2}
    \min_{\theta_{\rm em}, \theta_{\rm re}}
    &\IEEEeqnarraymulticol{3}{l}{ \alpha \mathcal{L_\text{R}} + \mathcal{L_\text{S}}},
\end{IEEEeqnarray}
where $\alpha \geq 0$ is a hyperparameter.

The embedding and recovery networks are trained to minimize $\mathcal{L_\text{R}}$ and $\mathcal{L_\text{S}}$ which yields to following optimization
\begin{IEEEeqnarray}{c'l'l'l}
    \IEEEyesnumber\label{timegan_Problem3}
    \min_{\theta_{g}}
    &\IEEEeqnarraymulticol{3}{l}{ \eta \mathcal{L_\text{S}} + \max_{\theta_{d}}\mathcal{L_\text{U}}},
\end{IEEEeqnarray}
where $\eta \geq 0$ is a hyperparameter.

In our training, we have implemented the TimeGAN in Tensorflow 2.0 \cite{abadi2016tensorflow} using original implementation in \cite{timegan-code}, and \textit{ydata-synthetic} package in \cite{ydata}. The hyper-parameters are listed in Table \ref{tab:timegan}.

\begin{table}
\caption{TimeGAN model and Hyper-parameters.}
    \centering
    \begin{tabular}{|c|c|}
    \hline
    \textbf{Hyper-parameters} & \textbf{Values}  \\
    \hline
    Sequence length & 600 \\
    \hline
    Number of features & 2 \\
    \hline
    Hidden units for generator & 24 \\
    \hline
    Gamma (used for discriminator loss) & 1 \\
    \hline
    Noise dimension (used by generator as a starter dimension) & 32 \\
    \hline
    Number of layers & 128\\
    \hline
    Batch size & 128 \\
    \hline
    Learning rate & $5$x$10^{-4}$\\
    \hline
    \end{tabular}
    
    \label{tab:timegan}
\end{table}

\subsection{Prediction Algorithms}
We design trajectory prediction algorithms with three different types of neural networks(NNs): Recurrent Neural Networks (RNN) \cite{giles1994dynamic}, Long Short Term Memory (LSTM) networks \cite{hochreiter1997long}, and Convolutional Neural Networks (CNN) \cite{gu2018recent}.

\subsubsection{RNN for Prediction} In the $t$-th time slot, the input of a RNN cell includes the feature observed in the current slot $\textbf{k}_t$, and the hidden state generated by the previous RNN cell, $\textbf{h}_{t-1}$. Then, RNN model updates the output and hidden states from the following steps,
\begin{align}
    \textbf{o}_t = \sigma(\textbf{W}_{\rm o}[\textbf{h}_{t-1},\textbf{k}_t] + \textbf{b}_{\rm o} )\;\text{and}\;
    \textbf{h}_t = \sigma(\textbf{W}_{\rm h} \textbf{o}_t + \textbf{b}_{\rm h}),\nonumber
\end{align}
where $\textbf{o}_t$ is the output of the RNN cell, $\textbf{h}_t$ is the new hidden state, $\textbf{W}_{\rm o}$ and $\textbf{W}_{\rm h}$ are weight matrices, $\textbf{b}_{\rm o}$ and $\textbf{b}_{\rm h}$ are bias terms, and $\sigma(\cdot)$ is the activation function.

\subsubsection{LSTM for Prediction} Each LSTM cell takes three inputs at each time slot: the feature observed in the current slot $\textbf{k}_t$, the previous LSTM cell state (i.e., long-term memory unit) $\textbf{L}_{t-1}$, and the previous hidden state (i.e., the short-term memory unit) $\textbf{h}_{t-1}$. Then, LSTM model updates the output, hidden state, and cell state as follows.
\begin{align}
    &\textbf{f}_t = \sigma(\textbf{W}_{\rm f}[\textbf{h}_{t-1},\textbf{k}_t] + \textbf{b}_{\rm f},\label{eq:LSTM1}\\
    &\textbf{i}_t = \sigma(\textbf{W}_{\rm i}[\textbf{h}_{t-1},\textbf{k}_t] + \textbf{b}_{\rm i}),\label{eq:LSTM2}\\
    &\textbf{\~{L}}_t = \tanh(\textbf{W}_{\rm k}[\textbf{h}_{t-1},\textbf{k}_t] + \textbf{b}_{\rm k}),\label{eq:LSTM3}\\
    &\textbf{L}_t = \textbf{f}_t \textbf{L}_{t-1} + \textbf{i}_t \textbf{\~{L}}_t,\label{eq:LSTM4}\\
    &\textbf{o}_t = \sigma(\textbf{W}_{\rm o}[\textbf{h}_{t-1},\textbf{k}_t] + \textbf{b}_{\rm o}),\label{eq:LSTM5}\\
    &\textbf{h}_t = \textbf{o}_t \tanh(\textbf{L}_t),\label{eq:LSTM6}
\end{align}

\noindent where $\textbf{f}_t$ is the forget gate which decides what information will be kept from the previous cell state, $\textbf{i}_t$ is the input gate which decides what information will be added to cell state of the network, $\textbf{o}_t$ is the output gate of the LSTM unit, $\textbf{h}_t$ is the new hidden state of the network, $\textbf{W}_f, \textbf{W}_i, \textbf{W}_k$, $\textbf{W}_o$ are the coefficient matrices, $\textbf{b}_f, \textbf{b}_i, \textbf{b}_k, \textbf{b}_o$ are the bias terms, and $\sigma(\cdot)$ is the activation function.

\subsubsection{CNN for Prediction}
\begin{figure}
    \centering
    \includegraphics[width = 0.49\textwidth]{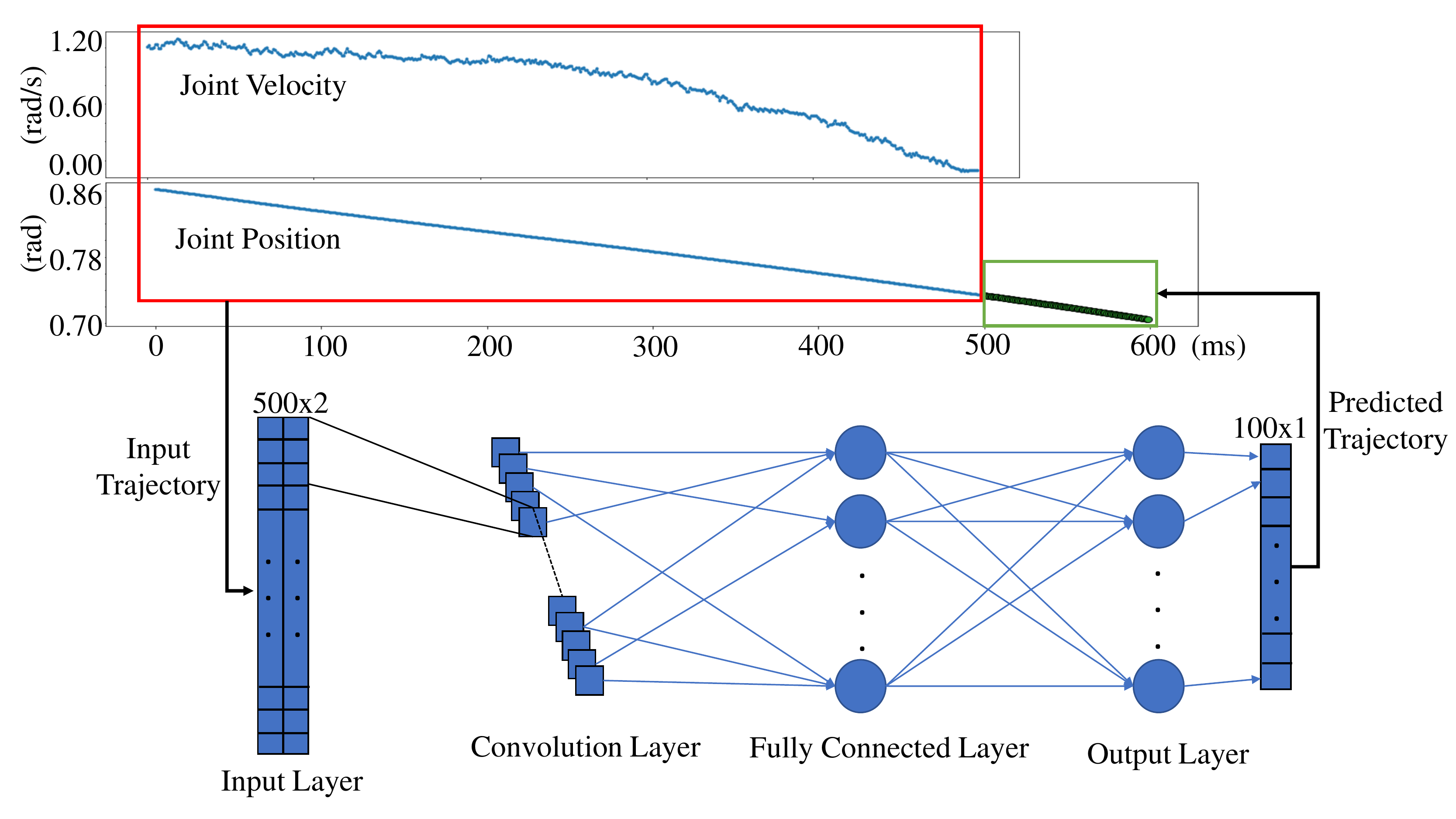}
    \caption{Illustration of trajectory prediction via CNN.}
    \label{fig:cnn}
\end{figure}

CNN consists of convolution layer, pooling layer and fully connected layer. In convolution layer the feature representations of inputs (i.e., \textit{feature maps}) are computed by applying element-wise convolution to the input with the kernel and then applying non-linear activation function to obtain the output of the layer. For input $\textbf{k}_t$ and kernel $\psi_t$, the resulting feature at location $(i,j)$ can be computed from the following steps,
\begin{align}
&\textbf{Z}_t^{i,j} = \textbf{W}_{\psi_t} * \textbf{k}_t^{i,j} + \textbf{b}_{\psi_t},\\
&\textbf{Y}_t^{i,j} = \Phi(\textbf{Z}_t^{i,j}),
\end{align}

\noindent where $\textbf{W}_{\psi_t}$ and $\textbf{b}_{\psi_t}$ are the weights and bias of the filter $\psi_t$, $\textbf{k}_t^{i,j}$ is the subsection of the input centered at $(i,j)$, $\Phi(\cdot)$ is the non-linear activation function, and `$*$' is the convolution operator. Then, the pooling layer is employed to decrease the number of features or resolution of the feature map after the convolution layer. The most used pooling operation is max-pooling \cite{boureau2010theoretical} which computes a new feature map by traversing the output of convolution layer and calculating the maximum of each patch (i.e., subsection of the convolution output according to filter size). After computing feature maps with several convolution and pooling layers, computed features are flatten to be fed to fully connected layer (i.e., dense layer) as seen in the Fig. \ref{fig:cnn} which is trained to forecast the future trajectories.

\begin{table*}
 \caption{Hyper-Parameters of Prediction Algorithms}
    \centering
    \begin{tabular}{|c|c|c|c|}
    \hline
        &\textbf{RNN} & \textbf{LSTM} & \textbf{CNN}\\
        \hline
        \textbf{Number of layers} & 1 & 1  & 1  \\
        \hline
        \textbf{Number of cells in each layer}&128 RNN cells & 128 LSTM cells & 128 Convolutional cells\\
        \hline
        \textbf{Batch size}& 64 & 64 &  64\\
        \hline
        \textbf{Optimizer}&Adam  & Adam  & Adam \\
        \hline
        \textbf{Loss function}&MSE  & MSE  & MSE \\
        \hline
        \textbf{Accuracy metric}&RRMSE & RRMSE & RRMSE \\
        \hline
        \textbf{Max training epochs}& 1000 &  1000 & 1000\\
        \hline
        \textbf{Early stopping criteria}& min validation loss, patience=10 &  min validation loss, patience=10 & min validation loss, patience=10\\
        \hline
         \textbf{Activation function}&tanh  & tanh  & ReLu \\
        \hline
    \end{tabular}
   
    \label{tab:pred_models}
\end{table*}

\subsection{Training and Testing of Prediction Algorithms}
For all the three types of NNs, we use mean squared error, $MSE = \sum_t(\hat{\textbf{k}}_t - \textbf{k}_t)^2 / s$, as the loss function since it is a differentiable function and eases the mathematical operations in optimizations throughout training process, where $s$ is the number of samples. In addition, we use relative root mean squared error, $RRMSE(\%) = \frac{\sqrt{\sum_t(\hat{\textbf{k}}_t - \textbf{k}_t)^2 / s)}}{\bar{\textbf{k}_t}} \times 100$, to compare the performance of different prediction algorithms, where $\bar{\textbf{k}_t}$ is the mean value of the observations. We set the maximum number of training epochs as $1000$. To avoid over fitting, an early stopping criteria is adopted. Specifically, the training process is terminated and the latest model is saved if the model does not improve for 10 consecutive epochs. The hyper-parameters of prediction algorithms are listed in Table \ref{tab:pred_models}.
To train the predictors, each input sample consists joint position and velocity in the past $N_{\rm in}$ time slots. Given the \textit{history window} with $N_{\rm in}$ number of steps, $\textbf{k}_{t-N_{\rm in}:t} = \{ \textbf{k}_{t-N_{\rm in}}, \textbf{k}_{t-N_{\rm in}+1}, ..., \textbf{k}_t \}$, predictor is trained to predict the joint positions in the next $N_{\rm out}$ number of steps (i.e., \textit{prediction window}) as shown in the example in Fig. \ref{fig:cnn}. In this study, we consider multi-step prediction in which predictor predicts $N_{\rm out}$ steps at once since it is more accurate than single-step prediction \cite{nikhil2018convolutional,zhang2018trajectory}.

The testing results are provided in Table \ref{tab:pred_accuracy}. Here, the dimension of the input is $N_{\rm in} = 500$~ms and the length of prediction window is $N_{\rm out} = 100$~ms. According to our results, CNN outperforms both LSTM and RNN. Therefore, we use CNN in our system and will be referring to CNN as the predictor in the rest part of the paper. It is worth noting that the prediction accuracies of RNN, LSTM, and CNN depend on the datasets \cite{huang2017study, khan2019electricity, li2020hybrid, yan2021multi}. In general, RNN and LSTM outperform CNN in time-series data. But for some datasets, where the time-series data change suddenly, CNN can be better than RNN and LSTM \cite{selvin2017stock}.

\begin{table}
\caption{Performance of Different Predictors}
    \centering
    \begin{tabular}{|c|c|c|c|}
    \hline
    \textbf{Errors(\%)}& \textbf{LSTM} & \textbf{RNN} & \textbf{CNN}  \\
    \hline
    \textbf{Training RRMSE} & $0.8\%$ & $0.5\%$ & $0.07\%$\\
    \hline
    \textbf{Validation RRMSE} & $0.8\%$ & $0.5\%$ & $0.08\%$\\
    \hline
    \textbf{Test RRMSE} & $0.9\%$ & $0.6\%$ & $0.2\%$\\
    \hline
    \end{tabular}
    \label{tab:pred_accuracy}
\end{table}

\subsection{Tradeoff between Prediction Error Probability and Prediction Horizon}
In this subsection, we illustrate how to obtain the tradeoff between the prediction error probability and the prediction horizon. The goal is the estimate the prediction error probability $f_{\epsilon^{\rm p}_m}(T_m^{\rm p},\delta_m)$, in \eqref{eq:perror}, i.e., the probability that the tracking error is greater than a required JND threshold $\delta_m$, when the prediction horizon is $T^{\rm p}_m$.

When the probability is extremely small, e.g., $f_{\epsilon^{\rm p}_m}(T_m^{\rm p},\delta_m) = 10^{-5}$, the real-world data set is not enough to obtain an accurate estimation. To overcome this difficulty, the synthetic trajectories generated by TimeGAN are used to estimate the prediction error probability. For every input trajectory $\textbf{k}_{t-N_{\rm in}:t} = \{k_{t-N_{\rm in}}, k_{t-N_{\rm in}+1},..., k_{t}\}$, there is a corresponding predicted trajectory $\textbf{\^{k}}_{t+1:t+N_{\rm out}} = \{\hat{k}_{t+1}, \hat{k}_{t+2},..., \hat{k}_{t+N_{\rm out}}\}$ and ground truth $\textbf{{k}}_{t+1:t+N_{\rm out}} = \{k_{t+1}, k_{t+2},..., k_{t+N_{\rm out}}\}$. The prediction errors with different prediction horizon (from $1$ time slot to $N_{\rm out}$ time slots) are given by $\textbf{e}_{t+1:t+N_{\rm out}} = \{e_{t+1}, e_{t+2},..., e_{t+N_{\rm out}}\}$, where $e_t = \hat{k}_{t} - k_{t}$. With both experimental data and synthetic data, we can model the prediction error probability with different prediction horizons and JND thresholds using feedforward neural network (FNN) that takes the prediction horizon, $T_m^{\rm p}$, and the JND threshold, $\delta_m$, as its inputs and outputs the prediction error probability, i.e.,
\begin{align}\label{eq:NumericalTradeoff}
   f_{\epsilon^{\rm p}_m}(T_m^{\rm p},\delta_m) = FNN(T_m^{\rm p},\delta_m|\phi_m),
\end{align}
where $\phi_m$ is the training parameters of the FNN. Hyper-parameters and training results are provided in Table \ref{tab:FNN}.

\begin{table}
\caption{Hyper-Parameters and Training Results of FNN}
    \centering
    \begin{tabular}{|c|c|}
    \hline
    \textbf{Number of layers} & 4  \\
    \hline
    \textbf{Number of cells in each layer} & 64 \\
    \hline
    \textbf{Batch size} & 64\\
    \hline
    \textbf{Activation} & ReLu\\
    \hline
    \textbf{Max training epochs} & 1000\\
    \hline
    \textbf{Loss function} & MSE\\
    \hline
    \textbf{Accuracy metric} & RRMSE\\
    \hline
    \textbf{Training RRMSE} & $0.6 \%$\\
    \hline
    \textbf{Validation RRMSE} & $0.9 \%$\\
    \hline
    \textbf{Test RRMSE} & $0.9 \%$\\
    \hline
    \end{tabular}
    \label{tab:FNN}
\end{table}

\section{Efficient Resource Allocation with Task-oriented Prediction}\label{resource allocation}
Based on the communication system in Section \ref{tradeoff_analyses} and the predictor in Section \ref{prediction models}, we illustrate how to optimize resource allocation, $W_m$, and data rate, $b_m$, in the task-oriented prediction and communication co-design framework. To maximize the number of users served by a BS, the optimization problem in \eqref{generalObjective}, can be re-formulated as follows,
\begin{IEEEeqnarray}{c'l'l'l}
    \IEEEyesnumber\label{timegan_Problem1}
    \max_{W_m, b_m}
    &\IEEEeqnarraymulticol{3}{l}{N} \label{Objective}\\
    \text{s.t.}\IEEEyessubnumber*
    & \sum_{m=1}^{N} W_m \leq W_{\max} & & \label{Const0}\\
    &  D^{\rm e}_m \leq D^{\max}_m & & \label{Const1}\\
    & \epsilon^\text{o}_m \leq \epsilon_{\max}  \label{Const2} \\
    & \eqref{overall_error}, \eqref{eps_d_formula}, \eqref{eps_q_formula}, \;\text{and}\; \eqref{eq:NumericalTradeoff}\nonumber,
\end{IEEEeqnarray}
where (\ref{Const0}) is the constraint on the maximum available bandwidth, (\ref{Const1}) and (\ref{Const2}) are the constraints on the QoS requirement. As shown in \eqref{overall_error}, the overall error probability is bounded by
\begin{align*}
        & \epsilon^{\rm ub}_m(W_m,b_m) \triangleq f_{\epsilon^\text{p}_m}(D_m^{\rm ch},\delta_m) \epsilon^\text{d}_m (1 - f_{\epsilon^\text{q}_m}(D_m^{\rm th})) \nonumber\\ 
        & + f_{\epsilon^\text{p}_m}(T_{\rm th},\delta_m) (f_{\epsilon_m^{\rm q}}(D_m^{\rm th}) - f_{\epsilon_m^{\rm q}}(D_m^{\rm th}+T_{\rm th})) \nonumber\\
        &+ f_{\epsilon^\text{q}_m}( D_m^{\rm th}+T_{\rm th}).\nonumber
\end{align*}
In the 5G New Radio system, the expression of the decoding error probability, $\epsilon^\text{d}_m$, is given by \eqref{eps_d_formula}. Given the queuing system in Section \ref{tradeoff_analyses}, the relationship between the queuing delay bound and the queuing delay violation probability, $f_{\epsilon^\text{q}_m}(\cdot)$, is derived in  \eqref{eps_q_formula}.
With our predictor in Section \ref{prediction models}, the tradeoff between the prediction error probability and the prediction horizon $f_{\epsilon^{\rm p}_m}(\cdot,\delta_m)$ is obtained in  \eqref{eq:NumericalTradeoff}. From \eqref{eps_d_formula}, \eqref{eps_q_formula}, and \eqref{eq:NumericalTradeoff}, we can see that the upper bound is determined by the bandwidth allocation and the data rate, and is denoted by $\epsilon^{\rm ub}_m(W_m,b_m)$. 

The number of constraints and the number of optimization variables are not deterministic in (\ref{Objective}), i.e., they depend on the number of users, $N$. Thus, it is hard to derive closed-form solution of this problem. To overcome this difficulty, we decompose the problem into multiple single-user subproblems.

\subsection{Single User Subproblem}
To maximize the number of users that can be served with a given bandwidth, we turn to minimize the bandwidth that is required to guarantee the QoS of each user. As such, the optimization problem can be decomposed into independent single-user subproblems, i.e.,
\begin{align}
    \min_{W_m, b_m}
    & \;{W_m} \label{Problem2}\\
    \text{s.t.}
    & \;\epsilon^{\rm ub}_m(W_m,b_m) \leq \epsilon_{\max}, \eqref{delay_req}, \eqref{overall_error}, \eqref{eps_d_formula}, \eqref{eps_q_formula}, \;\text{and}\; \eqref{eq:NumericalTradeoff}.\nonumber
\end{align}
We develop a two dimensional binary search algorithm to find the optimal solution of problem \eqref{Problem2}. We first fix the value of ${W_m}$ and find the optimal data rate that minimizes the upper bound of the overall reliability in \eqref{overall_error}, i.e.,
\begin{align}
\min_{b_m}&\;{\epsilon^{\rm ub}_m(W_m,b_m)} \label{Objective3}\\
\text{s.t.}&\;
\eqref{delay_req}, \eqref{eps_d_formula}, \eqref{eps_q_formula}, \;\text{and}\; \eqref{eq:NumericalTradeoff}\nonumber,
\end{align}
The optimal solution of the problem \eqref{Objective3} is denoted by $b_m^*(W_m)$. Then, we use binary search to find the value of $W_m$ that satisfies the following equation,
\begin{align}\label{eq:ReConstraint}
    \epsilon_m^{\rm ub}(W_m,b_m^*(W_m)) = \epsilon_{\max}.
\end{align}
We denote the solution of \eqref{eq:ReConstraint} by $W_m^*$. 

In the rest part of this section, we prove that $W^*_m$ and $b_m^*(W^*_m)$ are the optimal solution of problem \eqref{Problem2}. To prove this, we need two lemmas.

\textit{Lemma 3:} For given $W_m$, $\epsilon^{\rm ub}_m(W_m,b_m)$ first decreases and then increases with $b_m$.

We first validate that $\epsilon^{\rm ub}_m(W_m,b_m)$ first decreases and then increases with $b_m$ in two asymptotic scenarios: 1) $b_m$ is sufficiently small such that the decoding error probability is close to zero; 2) $b_m$ is sufficiently large such that the queuing delay violation probability is close to zero.

In the first scenario that $b_m$ is very small, the queuing delay violation probability $f_{\epsilon^\text{q}_m}(.)$ is large and, the decoding error probability $\epsilon^\text{d}_m$ is close to zero. Then, $\epsilon^{\rm ub}_m(W_m,b_m)$ can be simplified as $ f_{\epsilon^\text{p}_m}(T_{\rm th},\delta_m) (f_{\epsilon_m^{\rm q}}(D_m^{\rm th}) - f_{\epsilon_m^{\rm q}}(D_m^{\rm th}+T_{\rm th})) + f_{\epsilon^\text{q}_m}( D_m^{\rm th}+T_{\rm th})$. Since $f_{\epsilon^\text{q}_m}(.)$ is a decreasing function of $b_m$ according to Lemma 1, $\epsilon^{\rm ub}_m(W_m,b_m)$ decreases with $b_m$.

When $b_m$ is sufficiently large, the queuing delay violation probability $f_{\epsilon^\text{q}_m}(.)$ is close to zero, but the decoding error probability $\epsilon^\text{d}_m$ is large. Then, $\epsilon^{\rm ub}_m(W_m,b_m)$ is dominated by $\epsilon^\text{d}_m$, which increases with $b_m$ according to Lemma 2. In this case, $\epsilon^{\rm ub}_m(W_m,b_m)$ increases with $b_m$.

In non-asymptotic scenarios, it is difficult to prove Lemma 3. We will validate Lemma 3 with numerical results.

\textit{Lemma 4:} $\epsilon_m^{\rm ub}(W_m,b_m^*(W_m))$ decreases with $W_m$.

\textit{Proof:} From (\ref{overall_error}), (\ref{eps_q_formula}), and (\ref{eps_d_formula}), we can see that given the value of $b_m$, only the decoding error probability decreases with the value of $W_m$, and all the other terms remain constant. If $W_m<\tilde{W}_m$, then, $\epsilon_m^{\rm ub}(W_m,b_m^*(W_m)) > \epsilon_m^{\rm ub}(\tilde{W}_m,b_m^*(W_m))$. According to the definition of $b^*_m(\tilde{W}_m)$, it is the solution of \eqref{Problem2}. Given bandwidth $\tilde{W}_m$, the optimal value of $b_m$ that minimize the upper bound of the decoding error probability is $b^*_m(\tilde{W}_m)$. In other words, $\epsilon_m^{\rm ub}(\tilde{W}_m,b_m^*(W_m)) > \epsilon_m^{\rm ub}(\tilde{W}_m,b^*_m(\tilde{W}_m))$. Then, we have
$\epsilon_m^{\rm ub}(W_m,b^*_m(W_m)) > \epsilon_m^{\rm ub}(\tilde{W}_m,b^*_m(\tilde{W}_m))$.

Lemma 4 indicates that the optimal solution of problem (\ref{Problem2}) should satisfy (\ref{eq:ReConstraint}) and the optimal bandwidth can be obtained by binary search. To find the optimal $b_m$ for a given $W_m$, we can also use binary search as indicated by Lemma 3. Therefore, the optimal solution of problem (\ref{Problem2}) can be obtained via the two dimensional binary search given in Algorithm \ref{alg1}.

 \begin{algorithm}
 \caption{Algorithm for solving (\ref{Problem2})}
 \begin{algorithmic}[1]
 \renewcommand{\algorithmicrequire}{\textbf{Input:}}
 \renewcommand{\algorithmicensure}{\textbf{Output:}}
 \REQUIRE Delay requirement $D^{\max}_m$, reliability requirement $\epsilon_m^{\max}$, threshold for queuing delay $D^{\rm th}_m$, prediction horizon threshold $T_{\rm th}$, channel coherence time $D^{\rm ch}_m$, average packet arrival rate $\lambda_m$, JND threshold $\delta_m$, initial bandwidth $W_0$, maximum bandwidth $W_{\max}$, large-scale channel gain $\alpha_m$, small-scale channel gain $g_m$, transmit power $P_m$, single sided noise spectral density $N_0$, initial number of bits $b_0$, maximum number of bits $b_{\max}$.\\
 \ENSURE Optimal $W^{*}_m$, $b_m^*(W_m^*)$ to ensure URLLC QoS requirements for $m^{th}$ task.\\
 \textit{Initialisation} : $W_\text{L} = W_\text{0}$, $W_\text{R} = W_{\max}$, $W_{\rm mid} = (W_\text{L}+W_\text{R})/2$.
 \STATE Binary search $b_m$ in $[b_0,b_{\max}]$ and obtain $\epsilon_m^{\rm ub}(W_{\rm mid},b_m^*(W_{\rm mid}))$
  \WHILE{$|\epsilon_m^{\rm ub}(W_{\rm mid},b_m^*(W_{\rm mid})) - \epsilon_m^{\max}| > (\epsilon_m^{\max})^2$ \textbf{and} $W_\text{L} < W_\text{R}$}
  
  \IF{$\epsilon_m^{\rm ub}(W_{\rm mid},b_m^*(W_{\rm mid})) \leq \epsilon_m^{\max}$}
  \STATE $W^{*}_m = W_{\rm mid}$
  \STATE $W_{\rm R} = W_{\rm mid}$
  \STATE $W_{\rm mid} = (W_{\rm L}+W_{\rm R})/2$
  
  \ELSE
  \STATE $W^{*}_m = W_{\rm mid}$
  \STATE $W_{\rm L} = W_{\rm mid}$
  \STATE $W_{\rm mid} = (W_{\rm L}+W_{\rm R})/2$
  
  \ENDIF
  \STATE Binary search $b_m$ in $[b_0,b_{\max}]$ and obtain $\epsilon_m^{\rm ub}(W_{\rm mid},b_m^*(W_{\rm mid}))$
  \ENDWHILE

 \RETURN $W_m^*$, $b_m^*(W_m^*)$ 
 \end{algorithmic} 
 \label{alg1}
 \end{algorithm}

\section{Evaluation of the Proposed Framework}\label{evaluation}
In this section, we evaluate the proposed task-oriented prediction and communication co-design framework with numerical results where Table \ref{tab:params} provides the parameter settings. We consider the path loss model $10 \log_{10}(\alpha_m) = -128.1 -36.7 \log_{10}(d_m)$, where $d_m=200$ m is the distance between the BS and the receiver \cite{hou2019prediction}. The overall error probability, $\epsilon^o_m$, is obtained from (\ref{overall_error}).

\begin{table}
\caption{Numerical Values of Parameters for Overall Error \cite{hou2019prediction,hou2021intelligent}.}
    \centering
    \begin{tabular}{|c|c|}
    \hline
        \textbf{Parameter} & \textbf{Value}  \\
        \hline
        $\lambda_m$, average packet arrival rate & $100$ packets/s\\
        \hline
        $P_m$, maximum transmit power & $23$  dBm\\
        \hline
        $N_0$, single sided noise spectral density & $-144$  dBm/Hz\\
         \hline
         $D^{\rm r}_m$, core network and backhaul delay & $10$ ms \\
        \hline
        $D^{\rm t}_m$, transmission delay & $0.5$ ms\\
         \hline
        $D^{\rm ch}_m$, channel coherence time & $10$ ms \\
         \hline
        $T_{\rm th}$, prediction horizon threshold & $50$ ms\\
        \hline
        $D^{\rm th}_m$, queuing delay threshold & $\max\{D^{\max}_m- D^{\rm t}_m - D^{\rm r}_m,0\}$ \\
        \hline

    \end{tabular}
    
    \label{tab:params}
\end{table}

\begin{figure}
    \centering
    \includegraphics[width=0.45\textwidth]{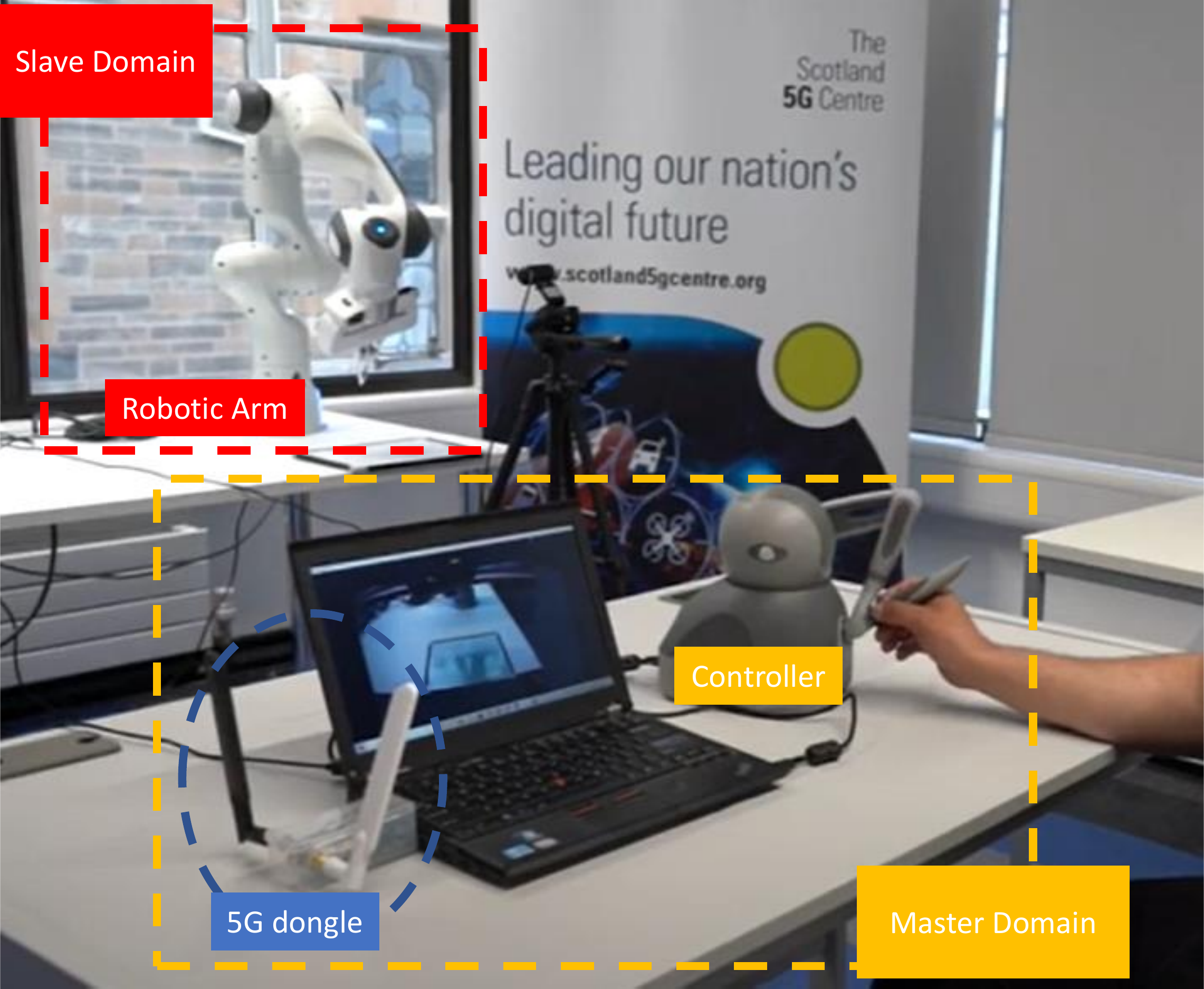}
    \caption{5G-enabled Teleoperation Prototype}
    \label{fig:prototype}
\end{figure}

We develop a 5G-enabled teleopeartion prototype to collect real-world trajectory data samples. As shown in Fig. \ref{fig:prototype}, a haptic device is deployed at the master domain as a controller, which has the capability of 6-degree-of-freedoms (DoFs) positional sensing and 3-DoF force feedback. We use a Franka Emika Panda robotic arm at slave domain, which is a 7-DoF serial manipulator with $1$~kHz control and sensor sampling capabilities. We perform mapping from controller's local coordinate system to robotic arm's local coordinate system since their DoFs and workspace are not identical. In the communication domain, we deploy a User Datagram Protocol (UDP) server at a 5G Base Station (BS) Mobile Edge Computing (MEC) unit. Both the controller and the robotic arm are equipped with 5G dongles which are small modems (i.e., 5G communication modules) built in-house at the University of Glasgow to access the local BS over 5G NR.

In the following, we first illustrate the tradeoff between prediction error probability and prediction horizon for different JND thresholds. Then, we consider the single-user scenario to illustrate the optimality of the Algorithm \ref{alg1} as well as we compare the two frameworks, where the predictors are either deployed at transmitter sides or at the receiver sides. Finally, we consider the multi-user scenario to evaluate the performance gain of the proposed framework in terms of the required bandwidth for each user or the maximum number of users that can be served.  

\subsection{Prediction Error Probability and Prediction Horizon Tradeoff}
\begin{figure}
    \centering
    \includegraphics[width=0.49\textwidth]{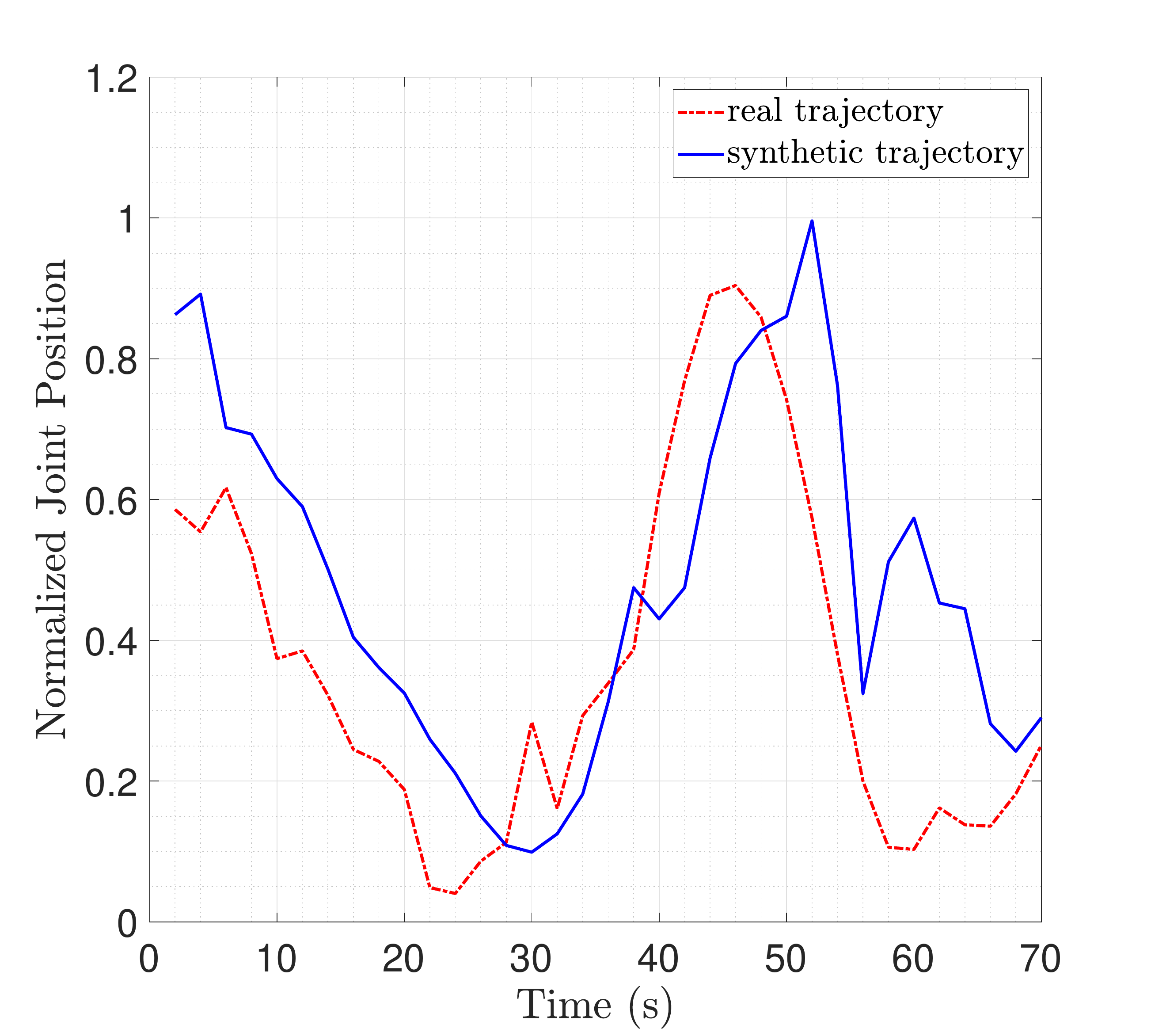}
    \caption{Real and synthetic trajectory comparison to illustrate the quality of generated data.}
    \label{fig:synt_real_comp}
\end{figure}

\begin{figure}
    \centering
    \includegraphics[width=0.49\textwidth]{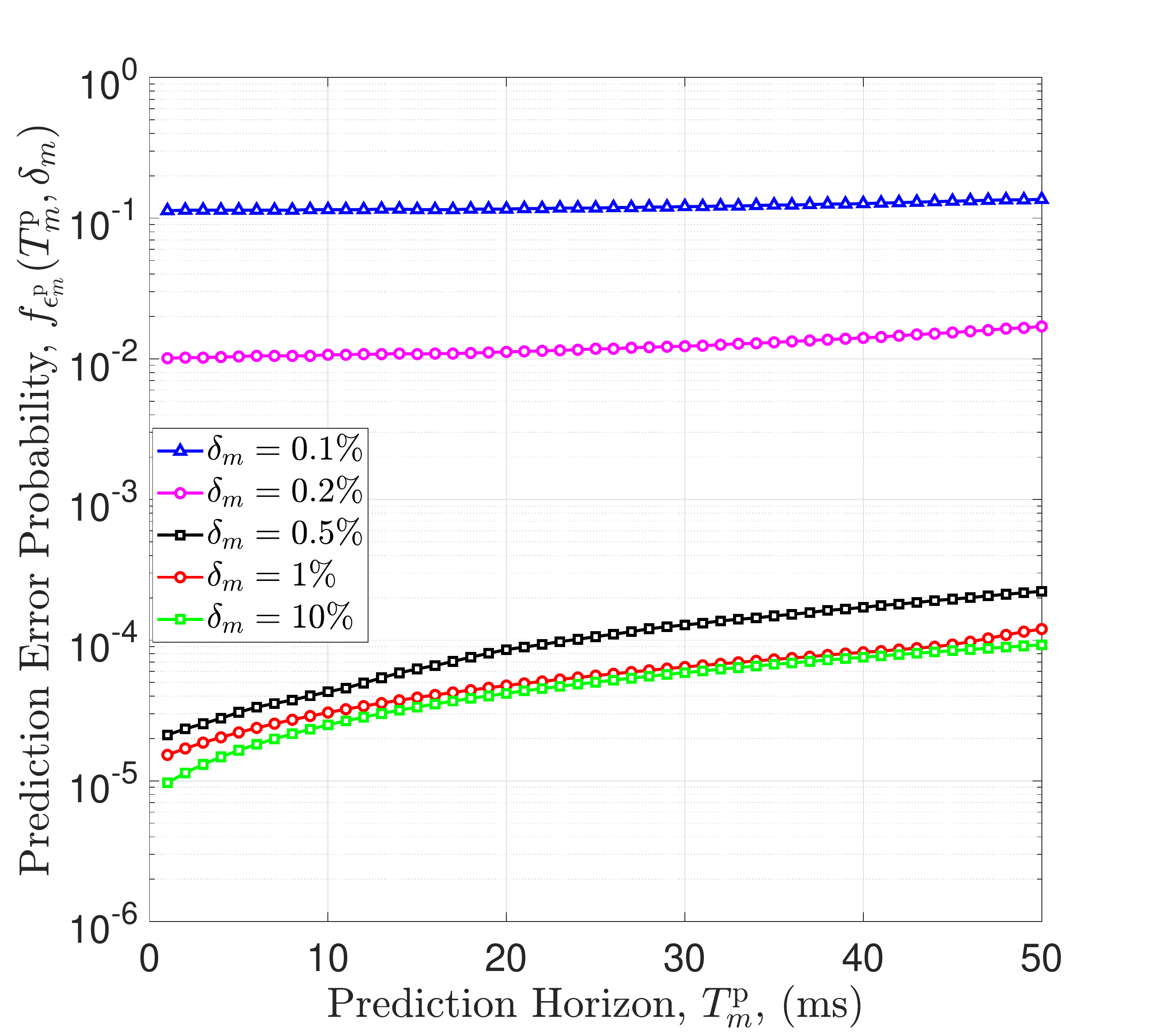}
    \caption{Prediction error probability versus prediction horizon for different JND thresholds. Prediction error probability curves for different JND thresholds are obtained from $1.7\times10^7$ real-world data samples and $2\times10^{11}$ synthetic data samples which correspond to $6.7\times10^8$ history and prediction window pairs.}
    \label{fig:tradeoff}
\end{figure}

In Fig. \ref{fig:synt_real_comp}, we compare a real-world trajectory segment with a synthetic trajectory segment. The results show that real and synthetic trajectories follow similar trend but with different values. This is reasonable since the motivation of generating synthetic data is for increasing the diversity of the dataset rather than generating the same data. Therefore, it is expected to have some differences between synthetic data and real data. We use both synthetic and real trajectories to obtain the tradeoff between the prediction error probability and the prediction horizon for different JND thresholds. In Fig. \ref{fig:tradeoff}, we provide the tradeoff between the prediction error probability and the prediction horizon. The results validate our assumption that the prediction error probability increases with the prediction horizon $T^p_m$ and decreases with the required JND threshold $\delta_m$. 

\subsection{Single-user Scenarios}
In single-user scenario, we consider two categories of tasks namely critical and non-critical tasks (see Fig. \ref{fig:tasks} where task 1 and task 2 are non-critical, and task 3 is critical), whose JND thresholds are $\delta_m=0.1\%$, and $\delta_m=1\%$, respectively. For fair comparison, we consider both tasks have the same value of delay requirement $D^{\max}_m$ as well as reliability requirement $\epsilon_m^{\max}$.

\begin{figure}
    \centering
    \includegraphics[width=0.49\textwidth]{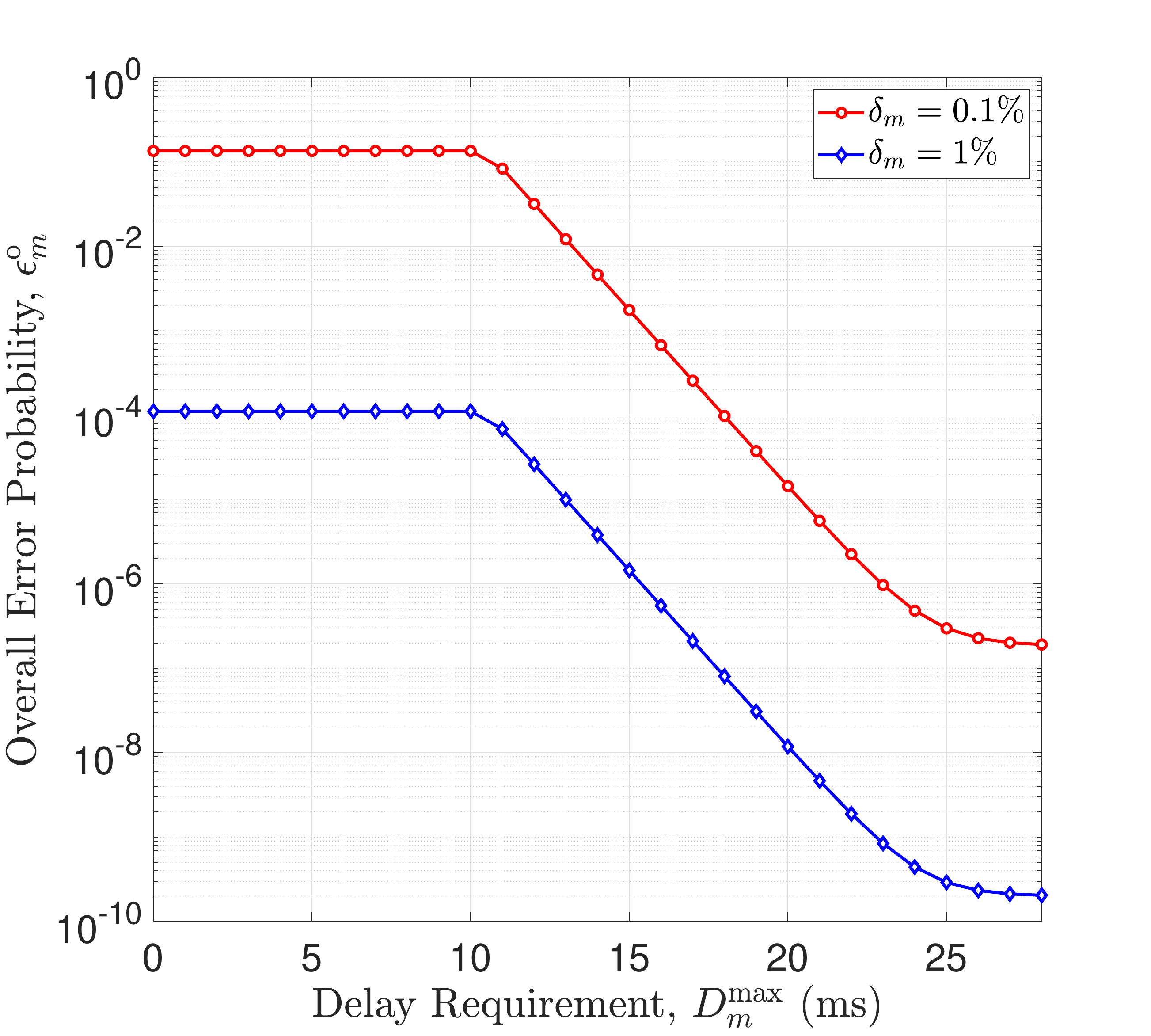}
    \caption{Overall error probability versus delay requirement, where $W_m = 140$ kHz and $b_m = 256$ bits.}
    \label{fig:overall_dmax}
\end{figure}

Fig. \ref{fig:overall_dmax} illustrates the trade-off between the required delay bound and the overall error probability with different JND thresholds. Here, the allocated bandwidth $W_m$ is $140 \text{ kHz}$ and the packet size $b_m$ is $256 \text{ bits}$. The results show that the overall error probability is constant and equal to prediction error probability when delay requirement smaller than or equal to the communication delay, i.e., $D_m^{\max} \leq D_m^{\rm c}$. Then, it decreases rapidly as the delay requirement $D_m^{\max}$ grows from $10$ to about $25$ ms. This is reasonable because the queuing delay violation probability $f_{\epsilon^\text{q}_m}(D^{\rm th}_m)$ is the dominant factor in the region $10< D_m^{\max} < 25 $ ms. When the delay requirement is larger than $25$ ms, both curves are nearly constant with $D_m^{\max}$. This is because $f_{\epsilon^\text{p}_m}(D^{\rm ch}_m, \delta_m) \epsilon^\text{d}_m$ becomes the dominant factor in (\ref{overall_error}) and does not change with the required delay bound.
Furthermore, the error probability of the critical task is much higher than that of the non-critical task. This implies that the bandwidth consumption of different task will be significantly different if they require the same error probability.

\begin{figure}
     \centering
     \begin{subfigure}[b]{0.49\textwidth}
         \centering
         \includegraphics[width=\textwidth]{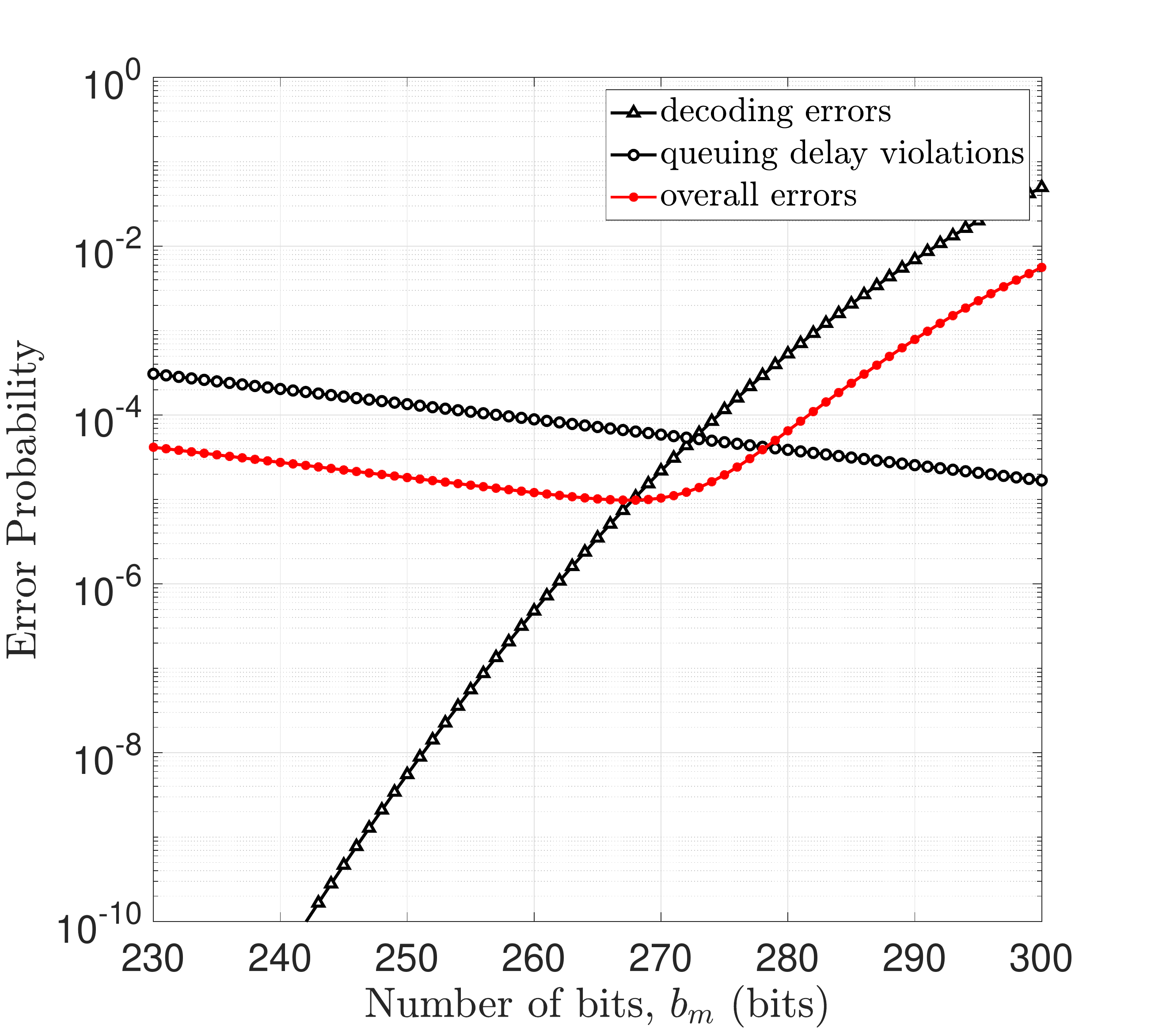}
         \caption{Error probability versus number of bits, where $\delta_m~=~0.1\%$.}
     \end{subfigure}
     \hfill
     \begin{subfigure}[b]{0.49\textwidth}
         \centering
         \includegraphics[width=\textwidth]{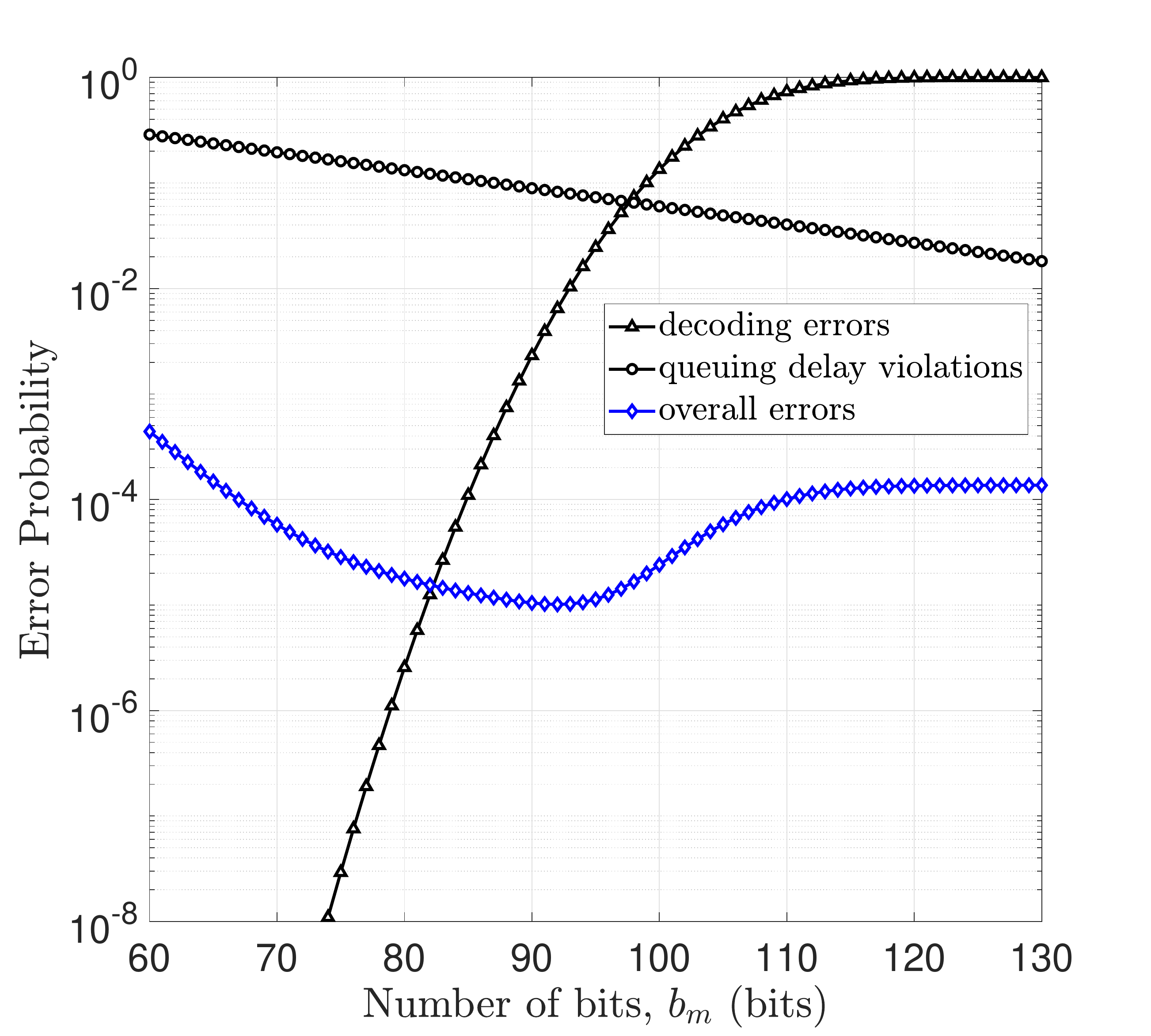}
         \caption{Error probability versus number of bits, where $\delta_m~=~1\%$.}
     \end{subfigure}
        \caption{Error probabilities versus number of bits, where $D^{\max}_m = 20$ ms and $\epsilon^{\max}_m = 10^{-5}$.}
        \label{fig:err_vs_bits}
\end{figure}

To validate Lemmas 1-3, we provide the relationship between different error probabilities in Fig. \ref{fig:err_vs_bits}. The results show that the queuing delay bound violation probability, $f_{\epsilon^\text{q}_m}(D^{\rm th}_m)$, decreases with number of bits and the decoding error probability, $\epsilon^\text{d}_m$, increases with number of bits. These two curves validate Lemma 1 and Lemma 2, respectively. For both critical and non-critical tasks, the overall error probability, $\epsilon^\text{o}_m,$ first decreases and then increases with number of bits. Moreover, when the queuing delay violations are larger than the decoding errors, the overall error is dominated by the queuing delay violation. When the decoding errors are larger than the queuing delay violations, the overall error is dominated by the decoding errors. These observations are consistent with Lemma 3. Furthermore, given the relationship between overall error, $\epsilon^{\rm o}_m$, and other error components such as decoding error probability, $\epsilon^\text{d}_m$, queuing delay violation probability, $f_{\epsilon^\text{q}_m}(.)$, and prediction error probability, $f_{\epsilon^\text{p}_m}(.,.)$, in equation (\ref{overall_error}) and given that $f_{\epsilon^\text{p}_m}(.,.) \leq 10^{-1}$, overall error probability could be lower than both decoding error and delay violations as seen in Fig. \ref{fig:err_vs_bits}.

\begin{figure}
    \centering
    \includegraphics[width=0.49\textwidth]{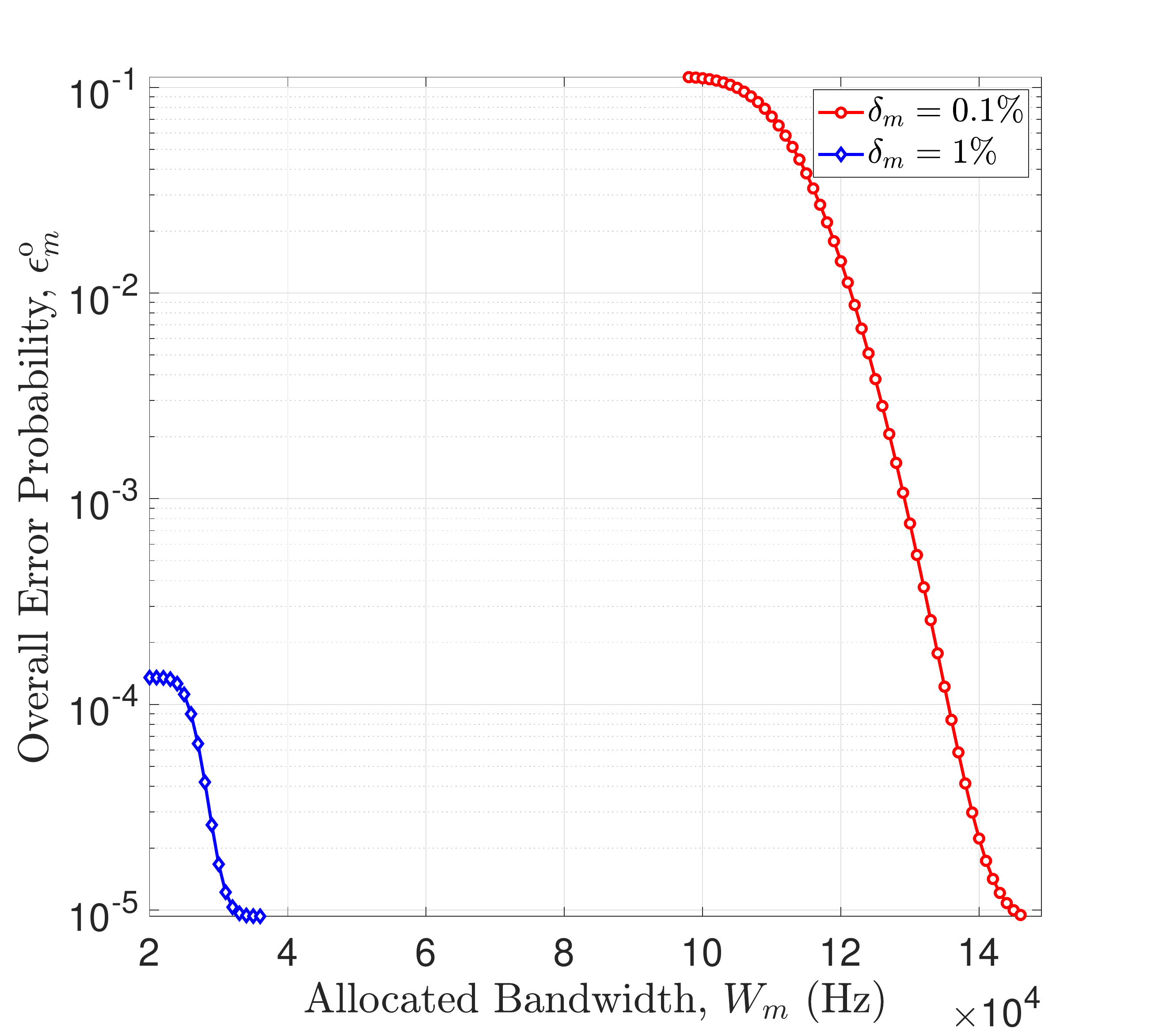}
    \caption{Error probabilities versus allocated bandwidth, where $D^{\max}_m = 20$ ms and $\epsilon^{\max}_m = 10^{-5}$.}
    \label{fig:err_vs_b}
\end{figure}

In Fig. \ref{fig:err_vs_b}, we compare error probabilities of critical and non-critical tasks under different values of allocated bandwidth, $W_m$. Here, the delay requirement is $20$~ms and the reliability requirement is $10^{-5}$. From the figure, the overall error probability decreases with increasing bandwidth in both critical and non-critical cases which is consistent with Lemma 4. The results in Figs. \ref{fig:err_vs_bits} and \ref{fig:err_vs_b} indicate that the overall error probability achieved by the proposed framework for critical case is $\epsilon_m^{\rm ub}(W^*_{m},b_m^*(W^*_{m})) = 1.00 \times 10^{-5}$ with $W^*_m = 145.24$ kHz, $b^*_m(W^*_m) = 268$ bits and for non-critical case is  $\epsilon_m^{\rm ub}(W^*_{m},b_m^*(W^*_{m})) = 1.00 \times 10^{-5}$ with  $W^*_m = 32.19$ kHz, $b^*_m(W^*_m) = 92$ bits.

In the existing literature, the predictor is either deployed at the transmitter \cite{hou2019prediction, tong2018minimizing, hosseini2016predictive, girgis2020predictive} or at the receiver \cite{boabang2020framework, girgis2021predictive}. Both deployment strategies have advantages and disadvantages. If the predictor is deployed at the transmitter, the advantage is that the historical information used in the prediction algorithm is accurate. The disadvantage is that either a prediction error or a packet loss in communication may result in a JND violation. If the predictor is deployed at the receiver, it can adjust the prediction horizon according to the communication delays of different packets. If the communication delay is satisfactory, there is no need to do any prediction. In this way, a JND violation happens when both the communication and the prediction fail. The disadvantage of this framework is that the historical information used in the prediction algorithm may not be accurate, because some packets are lost or severely delayed. Nevertheless, the effects of deployment strategy hasn't been investigated and deserve further analyses.

\begin{figure}
     \centering
     \begin{subfigure}[b]{0.49\textwidth}
         \centering
         \includegraphics[width=\textwidth]{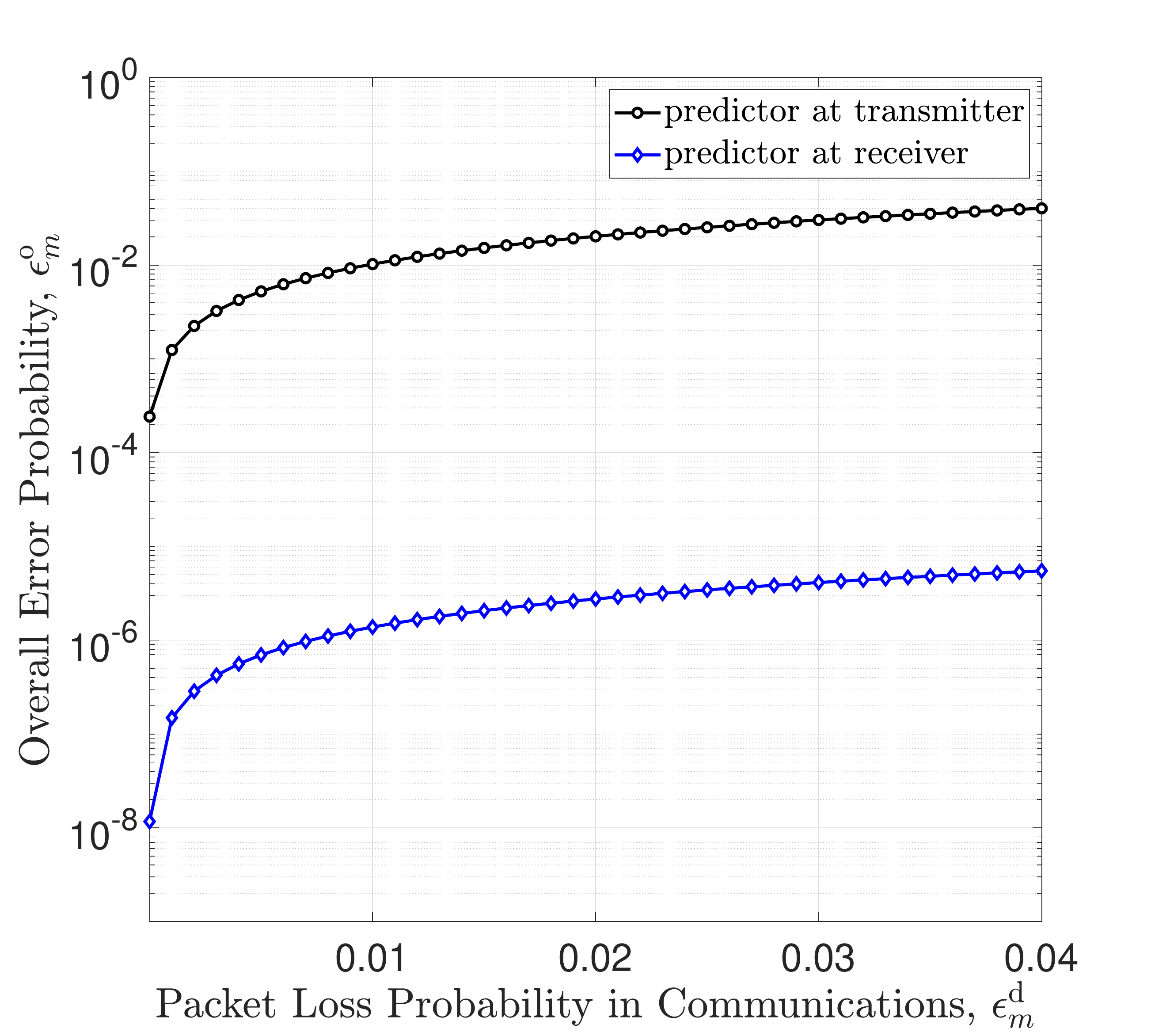}
         \caption{Overall error probability versus packet loss probability in communications, where $D^{\max}=20$~ms.}
     \end{subfigure}
     \hfill
     \begin{subfigure}[b]{0.49\textwidth}
         \centering
         \includegraphics[width=\textwidth]{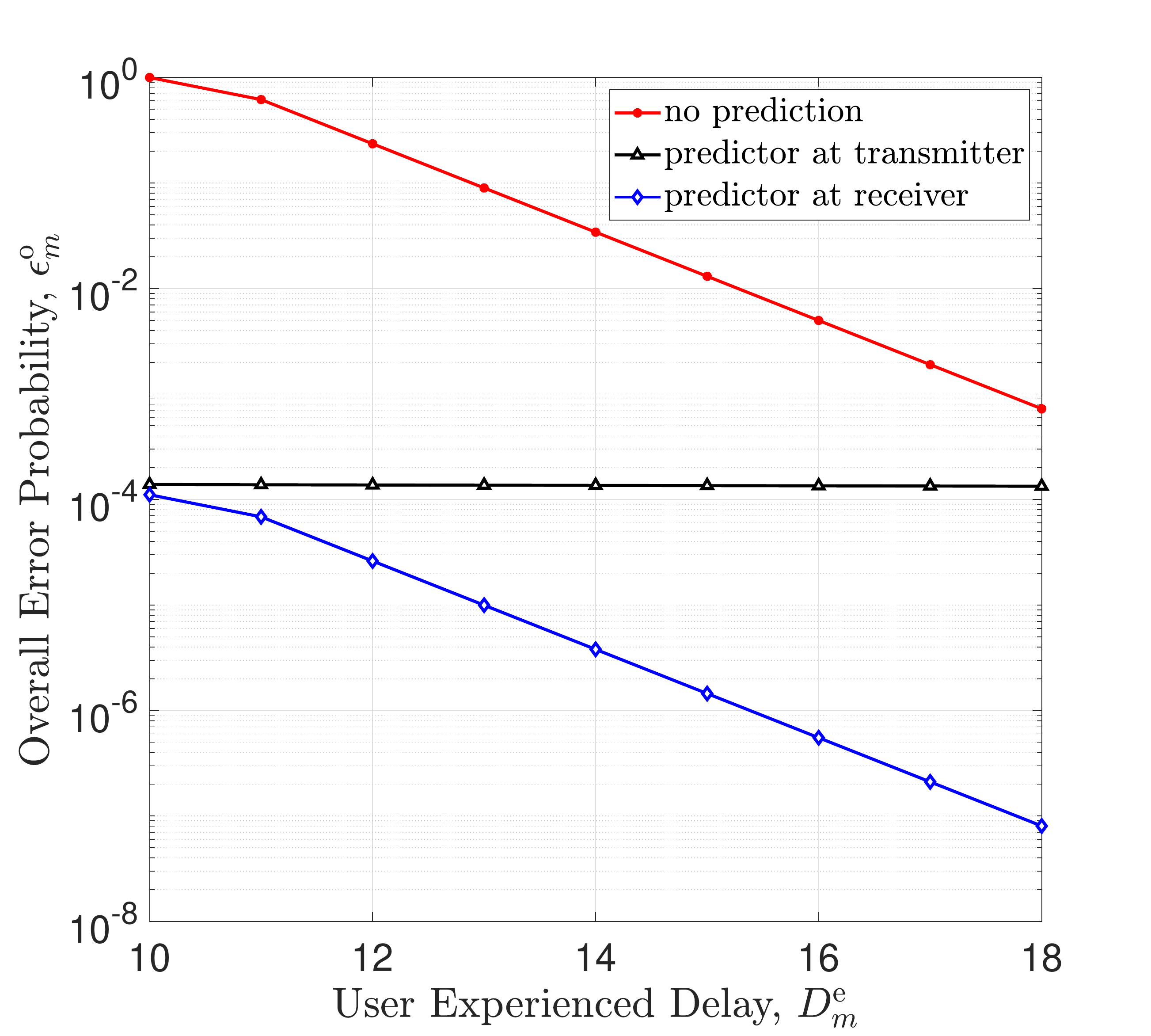}
         \caption{Overall error probability versus user experienced delay, where $\epsilon^\text{d}=10^{-5}$.}
     \end{subfigure}
        \caption{Predictor at transmitter versus predictor at receiver where, $W_m=140 $ kHz, $\delta_m=1\%$.}
        \label{fig:comparison}
\end{figure}

Here, we provide comparison between two deployment strategies to highlight the differences and deliver some insights for prediction and communication co-design research. Fig. \ref{fig:comparison} compares the overall system reliability of two deployment strategies under different values of packet loss probabilities in communications (Fig. \ref{fig:comparison}(a)) and user experienced delay (Fig. \ref{fig:comparison}(b)). From Fig. \ref{fig:comparison}(a), deploying predictor at receiver achieves better overall reliability with identical communication conditions. The reason behind this result is that if the predictor at the transmitter, then the overall system reliability dominated with least reliable system (i.e. either prediction system or communication system) since predicted future trajectories transmitted over communication system. However, if predictor at receiver, then it becomes compensation mechanism for communication system where lost packets can be predicted with cost of prediction errors. From Fig. \ref{fig:comparison}(b), deploying predictor at receiver can achieve better delay and reliability tradeoff. Both strategies can compensate or reduce user experienced delay. However, deploying predictor at receiver can achieve similar user experienced delay with higher reliability. The insight is that if the communication system reliability is high, i.e., packet loss probability less than $10^{-5}$, both strategies are suitable to reduce user experienced delay with accurate predictor. However, when the communication system is not reliable, i.e. packet loss probability greater than $10^{-5}$, we can achieve the URLLC QoS requirements by only deploying the predictor at the receiver.

\subsection{Multi-users Scenarios}
In multi-user scenarios, we compare the proposed task-oriented prediction and communication framework with the task-agnostic prediction and communication benchmark. Similar to single-user scenario, we assume two types of tasks and used the results from single-user scenario for the minimum required bandwidth and optimal packet rate. Furthermore, we denote critical task ratio to total number of tasks as $r=\frac{A}{A+B}, A\geq0,B\geq0$ where $A$ is the number of critical tasks and $B$ is the number of non-critical tasks.

\begin{figure}
     \centering
     \begin{subfigure}[b]{0.49\textwidth}
         \centering
         \includegraphics[width=\textwidth]{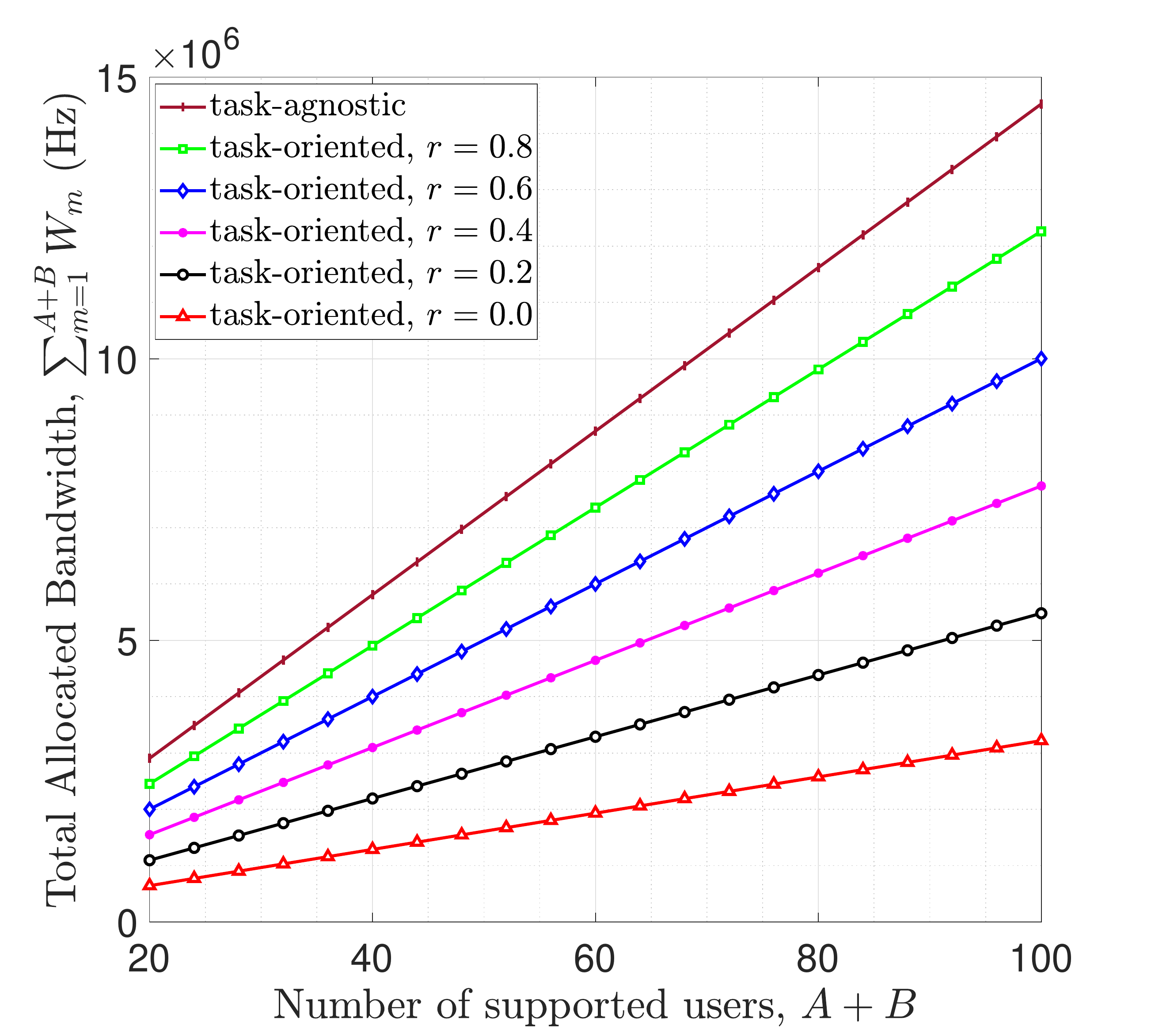}
         \caption{Total required bandwidth versus number of users.}
     \end{subfigure}
     \hfill
     \begin{subfigure}[b]{0.49\textwidth}
         \centering
         \includegraphics[width=\textwidth]{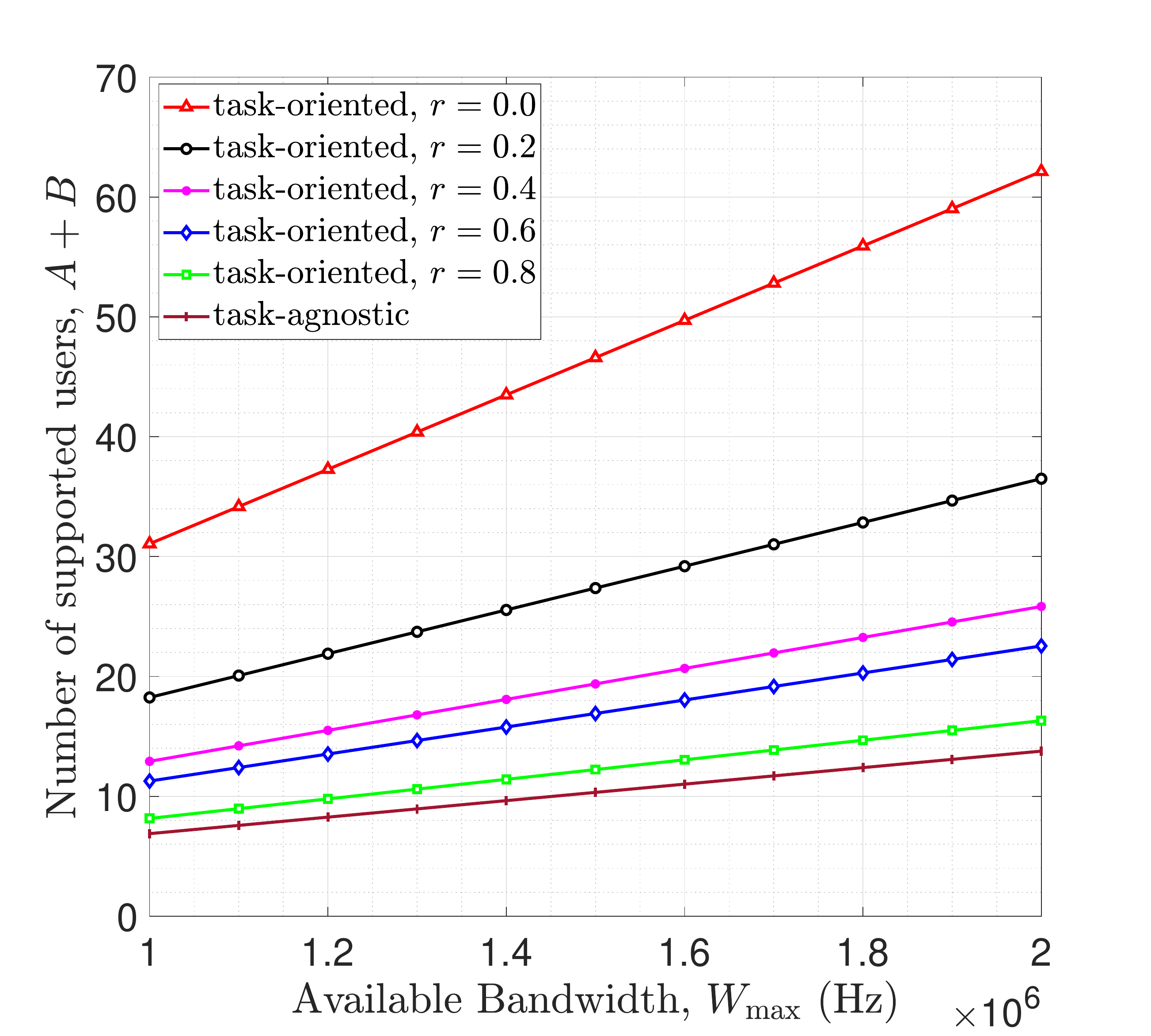}
         \caption{Number of supported users versus maximum available bandwidth.}
     \end{subfigure}
        \caption{The proposed task-oriented design versus task-agnostic design in terms of resource utilization.}
        \label{fig:user_vs_band}
\end{figure}
Fig. \ref{fig:user_vs_band} compares the resource utilization of the proposed task-oriented prediction and communication framework and the benchmark that is task-agnostic under different values of available resources and critical task ratios. From the figures, the proposed approach achieves more efficient use of resources with up to $77.80\%$ resource saving. This is because the proposed approach allocates resources according to JND thresholds of different tasks.

\section{Conclusions}\label{conclusion}
In this paper, we proposed the task-oriented prediction and communication co-design framework to increase the wireless resource utilization efficiency for haptic communications, where low latency and high reliability performance are required. The basic tradeoff between resource utilization efficiency and overall reliability are provided. For predictions, we see that real-world data are not enough to achieve URLLC level reliability. To address this issue, we deploy TimeGAN to generate realistic synthetic data which we use to reveal experimental tradeoff between prediction reliability and prediction horizon for different JND thresholds. We analysed prediction and communication systems and studied their relationship to reveal the tradeoff between wireless resources and reliability in proposed task-oriented communication and prediction co-design. We considered the teleoperation scenario via 5G New Radio and demonstrated a design example using the proposed framework. We formulated a joint optimization problem to maximize the number of users in a communication system by jointly optimizing communication data rate and task-dependent JND threshold. Numerical results show that the proposed approach can reduce the wireless resource consumption by $77.80\%$ compared with the benchmark that is task-agnostic.

As a future work, the proposed framework can be extended to a two-way prediction and communication framework, in which both transmitter and receiver can be equipped with predictors to predict both lost and delayed packets as well as network conditions to dynamically allocate resources accordingly. This sequential decision making mechanism can be achieved by training Deep Reinforcement Learning (DRL) agent with expert knowledge and real-time information available such as channel state information (CSI).

\bibliographystyle{IEEEtran}
\bibliography{ref}

\end{document}